%% file: paper.tex
\documentclass[journal,compsoc]{IEEEtran}

\usepackage{algorithmic}
\usepackage[]{algorithm2e}
\usepackage{booktabs}
\usepackage{adjustbox}
\usepackage{amsmath,mathtools}
\usepackage{graphics}
\usepackage{multirow}
\usepackage{xcolor}
\usepackage{subfig}
\usepackage{graphicx}
\usepackage{hhline}
\usepackage{makecell}
\usepackage{amssymb}
\usepackage{hyperref}
\usepackage{arydshln}
\usepackage{colortbl}
\usepackage{enumitem}
\usepackage{relsize}
\usepackage{courier}
\usepackage{xspace}
\usepackage{tikz}
\usepackage[utf8]{inputenc}

\usepackage[noadjust]{cite}
\usepackage[super]{nth}

\newcommand{\msn}{\texttt{MSN30K}\xspace}
\newcommand{\istella}{\texttt{Istella-S}\xspace}
\newcolumntype{R}[1]{>{\raggedleft\let\newline\\\arraybackslash\hspace{0pt}}p{#1}}

\AtBeginDocument{%
  \providecommand\BibTeX{{%
    \normalfont B\kern-0.5em{\scshape i\kern-0.25em b}\kern-0.8em\TeX}}}

\makeatletter
\def\adl@drawiv#1#2#3{%
        \hskip.5\tabcolsep
        \xleaders#3{#2.5\@tempdimb #1{1}#2.5\@tempdimb}%
                #2\z@ plus1fil minus1fil\relax
        \hskip.5\tabcolsep}
\newcommand{\cdashlinelr}[1]{%
  \noalign{\vskip\aboverulesep
           \global\let\@dashdrawstore\adl@draw
           \global\let\adl@draw\adl@drawiv}
  \cdashline{#1}
  \noalign{\global\let\adl@draw\@dashdrawstore
           \vskip\belowrulesep}}
\makeatother

\begin{document}
\title{Distilled Neural Networks for\\Efficient Learning to Rank}
\author{Franco Maria Nardini, Cosimo Rulli, Salvatore Trani, and Rossano Venturini
\IEEEcompsocitemizethanks{
\IEEEcompsocthanksitem Cosimo Rulli and Rossano Venturini are with the University of Pisa, Italy. E-mail: cosimo.rulli@phd.unipi.it, rossano.venturini@di.unipi.it.

\IEEEcompsocthanksitem  Franco Maria Nardini, Cosimo Rulli, Salvatore Trani and Rossano Venturini are with the ISTI–CNR, Pisa, Italy. E-mail: \{cosimo.rulli, f.nardini, s.trani, rossano.venturini\}@isti.cnr.it.}
}



\IEEEtitleabstractindextext{

\input{00.abstract.tex}

\begin{IEEEkeywords}
Web search, learning-to-rank, neural networks, efficiency, distillation, pruning, matrix multiplication.
\end{IEEEkeywords}
}



\maketitle
\begin{tikzpicture}[remember picture,overlay]
\node[anchor=south,yshift=5pt] at (current page.south) {\fbox{\parbox{\dimexpr\textwidth-\fboxsep-\fboxrule\relax}{
  \footnotesize{
     \copyright 2022 IEEE. Personal use of this material is permitted.  Permission from IEEE must be obtained for all other uses, in any current or future media, including reprinting/republishing this material for advertising or promotional purposes, creating new collective works, for resale or redistribution to servers or lists, or reuse of any copyrighted component of this work in other works.
  }
}}};
\end{tikzpicture}
\IEEEdisplaynontitleabstractindextext
\IEEEpeerreviewmaketitle

\vspace{-1.2cm}
\input{01.intro.tex}

\input{02.related.tex}
\input{03.methodology.tex}
\input{04.experiments.tex}
\input{05.conclusions.tex}

\bibliographystyle{IEEEtranS}
\bibliography{IEEEabrv,biblio}

\begin{IEEEbiography}[{\includegraphics[width=1in,height=1.25in,clip,keepaspectratio]{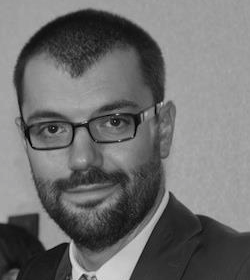}}]{Franco Maria Nardini}
is a senior researcher with the National Research Council of Italy. His research interests focus on web information retrieval and machine/deep learning.
He authored more than 70 papers in peer-reviewed international journal and conferences.
He received the Best Paper Award at ACM SIGIR 2015 and the Best Demo Paper Award at ECIR 2014.
For more information: \url{http://hpc.isti.cnr.it/\~nardini}.
\end{IEEEbiography}

\begin{IEEEbiography}[{\includegraphics[width=1in,height=1.25in,clip,keepaspectratio]{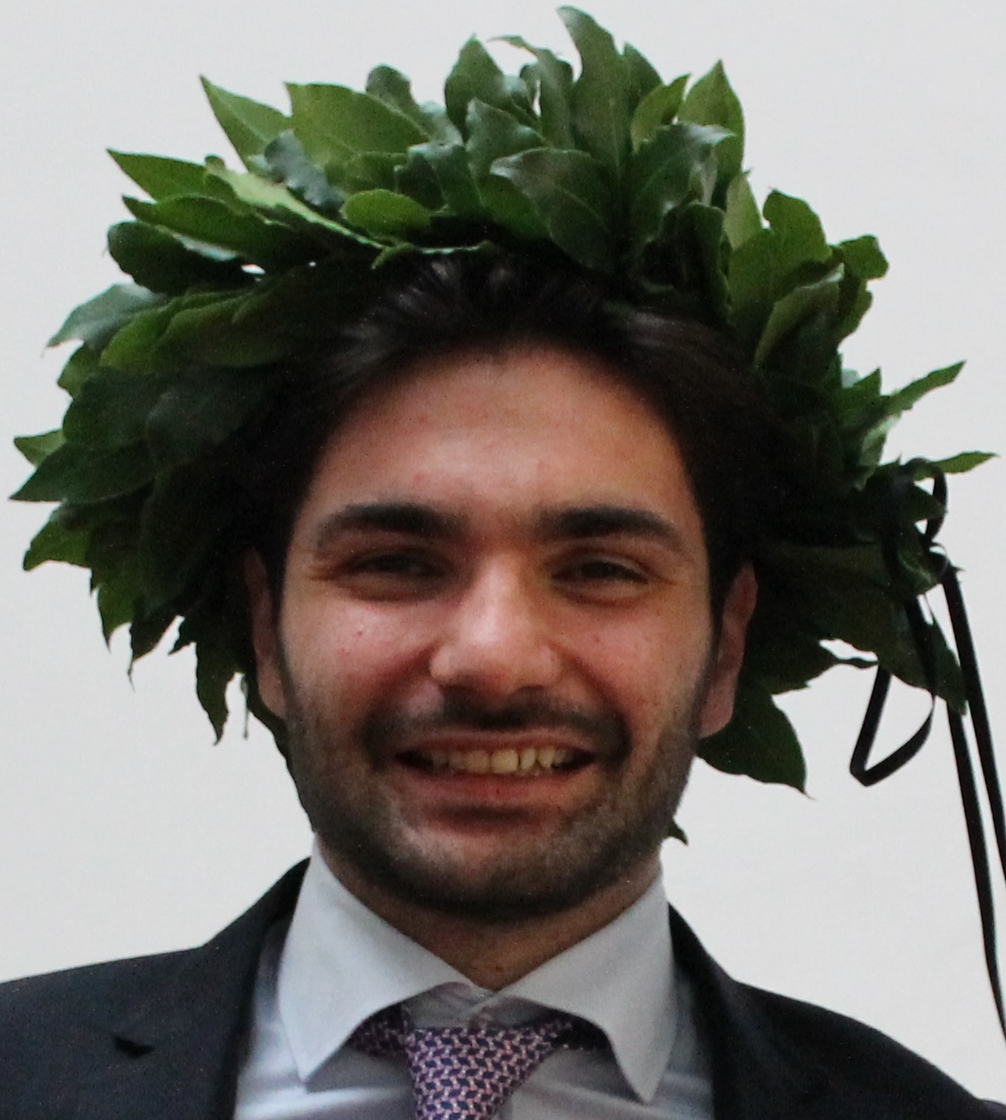}}]{Cosimo Rulli}
is a Ph.D student at the University of Pisa and a researcher with the National Research Council of Italy. He received the Master Degree from the University of Florence in 2019, with a thesis on deep neural network compression with knowledge distillation and pruning. His research interests focus on deep learning, model compression, and information retrieval. 
\end{IEEEbiography}

\begin{IEEEbiography}[{\includegraphics[width=1in,height=1.25in,clip,keepaspectratio]{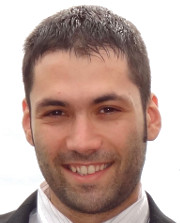}}]{Salvatore Trani}
is a researcher with the National Research Council of Italy. He received his PhD in Computer Science from the University of Pisa in 2017. His main research interests ranges from Information Retrieval to Web Mining and Machine Learning. He authored more than 15 papers on these topics, published in peer reviewed international journals and  conferences.
\end{IEEEbiography}

\begin{IEEEbiography}[{\includegraphics[width=1in,height=1.25in,clip,keepaspectratio]{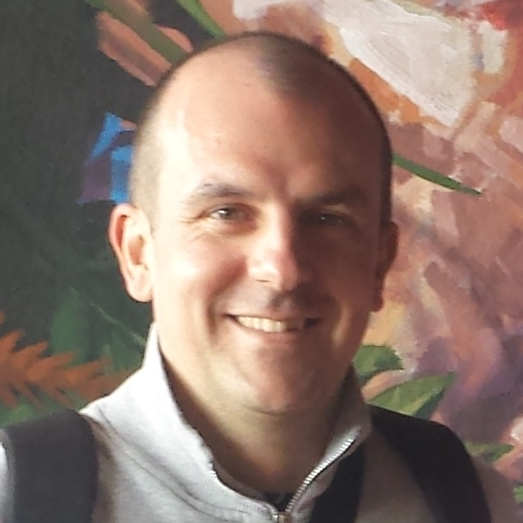}}]{Rossano Venturini}
is an associate professor at the Computer Science Department, University of Pisa. His research interests include the design and the analysis of algorithms and data structures with focus on indexing and searching large textual collections. He received two Best Paper Awards at ACM SIGIR in 2014 and 2015. For more information: \href{http://pages.di.unipi.it/rossano}{http://pages.di.unipi.it/rossano}.
\end{IEEEbiography}
\end{document}

%% file: 00.abstract.tex

\begin{abstract}
Recent studies in Learning to Rank have shown the possibility to effectively distill a neural network from an ensemble of regression trees. This result leads neural networks to become a natural competitor of tree-based ensembles on the ranking task. Nevertheless, ensembles of regression trees outperform neural models both in terms of efficiency and effectiveness, particularly when scoring on CPU. 
In this paper, we propose an approach for speeding up neural scoring time by applying a combination of Distillation, Pruning and Fast Matrix multiplication. We employ knowledge distillation to learn shallow neural networks from an ensemble of regression trees. Then, we exploit an efficiency-oriented pruning technique that performs a sparsification of the most computationally-intensive layers of the neural network that is then scored with optimized sparse matrix multiplication. Moreover, by studying both dense and sparse high performance matrix multiplication, we develop a scoring time prediction model which helps in devising neural network architectures that match the desired efficiency requirements. Comprehensive experiments on two public learning-to-rank datasets show that neural networks produced with our novel approach are competitive at any point of the effectiveness-efficiency trade-off when compared with tree-based ensembles, providing up to $4$x scoring time speed-up without affecting the ranking quality.
\end{abstract}

%% file: 01.intro.tex

\section{Introduction}
\label{sec:introduction}

\IEEEPARstart{T}{he} estimation of relevance is a task of paramount importance in Web search. In fact, search engines provide the users with a list of relevant results answering a information need formulated as a textual query. In the last years, Learning to Rank (LtR) techniques have been successfully applied to solve this task. LtR is the field of machine learning devoted to the development of supervised techniques addressing the ranking problem. LtR techniques have been proficiently used in Web search, a scenario characterized by tight latency bounds for query processing~\cite{cambazoglu2011scalability}. For this reason, the investigation of new LtR techniques targets both effectiveness and efficiency to provide accurate solutions that can be used in modern query processors. State-of-the-art approaches in learning to rank are ensembles of regression trees. Specifically, LambdaMART~\cite{burges2010ranknet} is an effective state-of-the-art LtR algorithm that builds ensembles of regression trees by optimizing a loss function that depends on a listwise information retrieval metric, e.g., NDCG~\cite{jarvelin2002cumulated}. The counterpart of the retrieval accuracy guaranteed by tree-based models is the computational effort needed to traverse hundreds or even thousands of trees. This computational effort hinders the application of this kind of models on low-latency query processors. Furthermore, each tree in an ensemble work by testing a sequence of boolean conditions on the input. The natural translation of this structure in \textit{if-then-else} code conflicts with modern CPU architectures that heavily rely on \textit{branch prediction} and \textit{caching}. A recent line of research investigates techniques for efficient traversal of ensembles of regression trees. The state-of-the-art algorithm for traversing tree-based models is QuickScorer~\cite{lucchese2015quickscorer,dato2016fast,8035185,lucchese2016exploiting}, which implements an interleaved feature-wise traversal of the ensemble that maximizes the efficiency of branch predictor and cache of modern CPUs.

Motivated by the success of neural solutions in other fields such as Natural Language Processing and Computer Vision, several attempts have been made to bring Neural Networks (NNs) in the LtR field. Despite that, tree-based solutions still provide state-of-the-art performances on different benchmarks, especially when dealing with handcrafted features~\cite{qin2020neural}. 
Recently, Qin \emph{et al}~\cite{qin2020neural} identify the reasons for the superiority of tree-based solutions in i) the sensitiveness of neural network to input features scale and transformations, ii) the lack of expressiveness in mostly adopted neural models in LtR, iii) the limited size of available LtR datasets w.r.t. to Natural Language Processing or Computer Vision. Cohen \textit{et al.}~\cite{cohen2018universal} develop an approach that permit to overcome these limitations on standard LtR datasets by training classic multi-layer perceptrons using simple data normalization ($Z$-normalization) and by leveraging a data augumentation technique (Section \ref{sec:cohen}). 
Cohen \textit{et al.}~\cite{cohen2018universal} propose to train neural networks to mimic the outputs of a pre-trained ensemble of regression trees. They do so by employing a knowledge distillation approach~\cite{ba2014deep,DBLP:journals/corr/HintonVD15} that treats the ensemble of regression trees as a black box generating accurate document scores.
Given that neural models are universal approximators~\cite{hornik1991approximation}, the network can reproduce the predictions of the ensemble of regression trees. In practice, this is done by using the Mean Square Error between the scores and the network predictions as training loss. The performance of a neural network trained by scores approximation are bounded by the performance of the tree-based model used to generate the scores: even in a perfect approximation scenario, the neural model will introduce no improvement in terms of effectiveness. In general, instead, the approximation will cause a degradation in the ranking precision. However, the reason to move to a neural document scoring engine is to exploit fast inference mechanisms available for NNs.
In this direction, Cohen \textit{et al.}~\cite{cohen2018universal} compare the efficiency of a neural solution for ranking (on GPU and CPU) with QuickScorer~\cite{lucchese2015quickscorer} (on CPU).
In the original work, the authors claim that neural models are as accurate as ensembles of regression trees in terms of Mean Average Precision (MAP), and largely outperform them in terms of execution time ($\mu$s/doc). We observe that their comparison presents some weaknesses.
They compare a single-thread CPU version of QuickScorer against a multi-thread GPU version of the neural forward pass. Due to the differences between the computational engines, this does not permit to actually state which one of the two solutions is the more efficient.
 Even when comparing on CPU, the comparison is done using: i) a single-threaded C++ implementation of QuickScorer for ensembles of regression trees and ii) a multi-threaded Python neural inference running with an unspecified number of threads. The use of Python APIs may also entail some latency in calling the underlying optimized matrix multiplication routine on which these frameworks usually rely.\footnote{See for example \url{https://scipy-cookbook.readthedocs.io/items/ParallelProgramming.html\#Use-parallel-primitives}} 
Moreover, the two sets of experiments are conducted on different CPUs. These aspects hamper a direct comparison of the performance achieved.

In this article, we propose a solid, fair and comprehensive comparison of the efficiency of ensemble of regression trees and neural models.
We compare QuickScorer~\cite{dato2016fast} against a novel and optimized implementation of neural network inference written in C++. We perform the evaluation on the same hardware by executing the two solutions using a single thread. Moreover, both solutions exploit instruction-level parallelism (AVX2 instruction set). Since CPU and GPU are two different processing units and each of them requires specific optimization techniques, in this work we focus on providing an accurate study of the efficiency of the two approaches on CPU, while we plan to extend it to the GPU in the future. Regarding the training phase, we adopt the same neural architectures of Cohen \textit{et al.}~\cite{cohen2018universal} and we re-implement their methodology with our own code in Pytorch~\cite{NEURIPS2019_9015}. However, differently from the original work, in our experiments we train the ensemble of regression trees with the LightGBM library~\cite{NIPS2017_6907}, since it is the state-of-the-art library for learning ensemble models on ranking tasks~\cite{NIPS2017_6907,qin2020neural}.

\begin{table}[htb]
\centering
\begin{tabular}{llllr}	
		\toprule
		Model & NDCG@10   & NDCG&  MAP & \thead{ Scoring Time \\ ($\mu s$/ doc)} \\
		\midrule
		Large Forest & 0.5246\textsuperscript{$\star \dag$} & 0.7473\textsuperscript{$\star \dag$} & 0.6604\textsuperscript{$\star \dag$} & 8.2  	\\
		\cdashlinelr{1-5}
		Mid Forest & 0.5206\textsuperscript{$\dag$}&0.7454\textsuperscript{$\dag$} &  0.6582\textsuperscript{$\dag$}& 1.5 	\\
		Small Forest & 0.5181& 0.7438 &  0.6578& 0.8 \\
		\midrule
		Large Net & 0.5198\textsuperscript{$\dag$} & 0.7445\textsuperscript{$\dag$} & 0.6582\textsuperscript{$\dag$} & 24.4 \\
		Small Net & 0.5171 &0.7432 & 0.6575   & 2.2 \\
		\bottomrule
\end{tabular}
\caption{A comparison between QuickScorer and Neural Networks on the \msn dataset. 
Symbols evidence statistically significant improvement w.r.t. to Mid Forest (\textsuperscript{$\star$}), and Small Forest (\textsuperscript{$\dag$}),  according to the Fisher's randomization test,  $p < 0.05$.}
\label{tab:firsttab}
\end{table}


The results of our comprehensive experimentation on the \msn dataset show that, in contrast with the results reported by Cohen \textit{et al.}~\cite{cohen2018universal}, ensembles of regression trees are both faster and more accurate than neural models. In Table \ref{tab:firsttab}, we report the Mean Average Precision (MAP), the Normalized Discounted Cumulative Gain (NDCG, with cutoff at 10 and without cutoff), and the scoring time per document.
Symbols evidence statistically significant improvement w.r.t. to Mid Forest\textsuperscript{$\star$}, and Small Forest\textsuperscript{$\dag$}, according to the Fisher's randomization test,  $p < 0.05$. We run different tests for each metrics, but we use shared symbols to ease the notation.
Table \ref{tab:firsttab} shows that ensemble of regression trees deliver the same performance of neural models while being largely faster, with a speedup ranging from $2.8$x (Small Net vs Small Forest) to $16.2$x (Large Net vs Mid Forest). Also, the Large Forest is the best performing model with a large margin, while being $3$x faster than the Large Net.
 These evidences highlight how tree-based solutions are currently faster than neural networks on CPU. We bridge the large gap between tree-based models and neural networks by proposing a novel framework to efficiently design and train effective and efficient feed-forward networks for ranking on CPU.

The novel contributions of this article are:
\begin{itemize}
\item we present a combination of state-of-the-art approaches to improve the performance of neural networks on Learning to Rank tasks. By leveraging efficiency-oriented pruning techniques and high-performance Dense and Sparse Matrix Multiplication techniques, we build neural models that outperform ensembles of regression trees. An extensive experimental evaluation on two well-established public benchmarks, \textit{i.e.}, the \msn~\cite{DBLP:journals/corr/QinL13} and the Tiscali \istella~\cite{dato2016fast} datasets, shows the effectiveness of our method. Experimental results confirm that on the \msn dataset it is possible to obtain up to $4.4$x faster scoring time with no loss of accuracy.

\item we provide a novel way to estimate the execution time of neural network forward pass, by mean of dense and sparse time predictors, respectively for Dense-Dense and Sparse-Dense Matrix Multiplication (DMM \& SDMM). To the best of our knowledge, this is the first work that dives into the technicality of matrix multiplication to precisely predict the execution time of neural models.
These predictors are derived from a broad study of the implementation of the relative operations on modern CPUs. In explaining how predictors are developed, we also provide a clear and concise explanation of these two fundamental operations with plenty of scientific applications. 

\item we develop an efficient and effective approach to design neural models, using the aforementioned time predictors, which allow to estimate the execution time of a feed-forward network \textit{a priori}, by providing the architecture - \textit{i.e.,} the number of layers and the neurons per layer - and the sparsity level of each layer. This design methodology tackles the costly problem of model architectures search~\cite{strubell2019energy,patterson2021carbon}, since it allows to train \emph{exclusively} the models respecting the latency requirements, tearing down the costs, in terms of time and energy consumption, of the experimental phase.
\end{itemize}

The rest of the paper is organized as follows: Section~\ref{sec:related} discusses the related work in the field. Section~\ref{sec:cohen} details the process of distilling ensemble of regression trees into neural networks as proposed by Cohen \emph{et al.}~\cite{cohen2018universal}. Section~\ref{sec:ModelMatMult} introduces the implementation of dense-dense matrix multiplication and sparse-dense matrix multiplication on modern CPUs, together with our time predictors. Section~\ref{sec:neuraleng} describes our novel method for designing efficient neural models for ranking. Moreover, Section~\ref{sec:experiments} presents a comprehensive experimental evaluation of our proposed technique on public data. Finally, Section~\ref{sec:conclusions} concludes the work.

%% file: 02.related.tex

\section{Related Work}
\label{sec:related}

In this section, we introduce Learning to Rank (LtR) and its use in Information Retrieval (IR). Then, we describe QuickScorer~\cite{lucchese2015quickscorer,dato2016fast,8035185} an efficient algorithm for scoring ensemble of regression trees. Finally, we discuss the field of model compression, a branch of machine learning that aims to compress Deep Neural Networks without affecting their accuracy. Here, we focus our attention in particular on pruning techniques.

\subsection{Learning to Rank}
\label{subsec:ltr}
Learning to Rank (LtR) consists in applying machine learning techniques to the problem of ranking documents with respect to a query.
RankNet~\cite{burges2005learning} leverages a  probabilistic ranking framework based on a pairwise approach to train a neural network. The difference between the predicted scores of two different documents is mapped to a probability by means of the sigmoid function. Hence, using the cross-entropy loss this probability is compared with the ground truth labels, and Stochastic Gradient Descent (SGD) is used to minimize this loss. FRank~\cite{tsai2007frank} exploits a generative additive model and substitutes the cross-entropy loss with the fidelity loss, a distance metric adopted in physics, superior to cross-entropy when applied on top of the aforementioned probabilistic framework since 1) has minimum in zero, 2) is bounded in $[0,1]$. 
Neither RankNet nor FRank directly optimize a ranking metric (\emph{e.g.}, NDCG), and this discrepancy weakens the power of the model. Since ranking metrics are flat and discontinuous, coding them into the loss function is troublesome.
To overcome this issue, LambdaRank~\cite{burges2007learning} heuristically corrects the RankNet gradients, exploiting the rank position of the document in the overall sorting: it multiplies the RankNet gradient with a term that measures the increase in terms of NDCG when switching the terms, generating the so-called $\lambda$-gradients.
McRank~\cite{li2008mcrank} casts the problem of ranking as MultiClass classification task, using a boosting tree algorithm to learn the class probabilities and then converting them into relevances with the expected relevance, outperforming LambdaRank.  This work also highlights that modeling the ranking problem as a classification task works better than modeling it as a regression one.  
LamdaMART~\cite{burges2010ranknet} combines the successful training methodology provided by $\lambda$-gradients with Multiple Additive Regression Trees (MART) - as McRank~\cite{li2008mcrank}, and it has been establishing as the state-of-the-art in LtR. Currently, ensembles of regression trees are the most effective solution among LtR techniques when dealing with handcrafted features. In the next section, we describe state-of-the-art approaches for efficient traversal of these trees, in order to employ them in latency-bound scenarios.

\subsection{Efficient Traversal of Tree-based Models}
\label{subsec:quickscorer}
QuicksScorer~\cite{lucchese2015quickscorer} is a state-of-the-art algorithm that allows to speedup the traversal of an ensemble of regression trees. As detailed in the previous section, ensemble of regression trees is the model exploited by several state-of-the-art learning-to-rank solutions, e.g., LambdaMART~\cite{burges2010ranknet}.
QuickScorer codes each tree of the ensemble as a bitvector of length $n$, where $n$ is the number of leaves, which is used to select the \textit{exit leaf} in the tree. Furthermore, each decision node in each tree is associated with a bitvector of the same length called \textit{mask}. If the corresponding test is evaluated to false, the bits corresponding to the unreachable leaves are set to zero. By performing the logical \texttt{AND} among all the masks, we obtain another bitvector, named \emph{leafidx}, in which the first one entry corresponds to the exit leaf. To efficiently compute the exit leaf, QuickScorer process all the nodes in a \textit{feature by feature} fashion. For each feature $f$, the associated thresholds among all the nodes in the forest are sorted in ascending order. Let us a consider a threshold $\gamma$ associated with a node $g$: when $x_f > \gamma$, 
the corresponding \emph{leafidx} is updated performing the \texttt{AND} operation with the \emph{mask} relative to $g$. Since the thresholds are sorted, as soon as $x_f \leq \gamma$, the evaluation of the current feature is interrupted, since the following instances will evaluate true as well. To further improve the efficiency of the algorithm, two variations of the original algorithm are introduced:  1) Block-Wise QuickScorer (BWQS), in which the forest is partitioned into blocks of trees fitting the L3 cache, reducing the cache-miss ration and 2) Vectorized QuickScorer (vQS)~\cite{lucchese2016exploiting}, in which scoring is vectorized using AVX2 instructions and 256-bit registers, allowing to process up to $8$ document at time. 
Lettich \emph{et al.}~\cite{8035185} propose a GPU version of QuickScorer, to exploit the massive parallelism of this computational engine. By properly managing the GPU memory hierarchy and furnishing an adequate degree of parallelism in the document scoring process, this version results up to 100x faster than the corresponding CPU version, when dealing with very large forests ($20$,$000$ trees).
 
The cost of traversing an ensemble of regression trees with QuickScorer depends on the number of false nodes, rather than on the length of the root-to-leaf paths. Since machine-learnt trees are imbalanced, the authors experimentally show that this reduces the percentage of nodes to evaluate from 80\% of classical traversal to the 30\% of QuickScorer~\cite{lucchese2015quickscorer}. Moreover, QuickScorer is implemented carefully taking into account cache and CPU issues. For example, QuickScorer structures are accessed sequentially thus favoring pre-fetching and avoiding branch mispredictions. However, when the number of leaves is larger than $64$, scoring a model with QuickScorer can be inefficient. Recently, RapidScorer tackles the problem of forest with a larger number of leaves~\cite{ye2018rapidscorer}. In fact, when $|\text{leaves}| > 64$, the logical \texttt{AND} between the bitvectors cannot be carried out in just one CPU instruction, hampering efficiency. For this reason, RapidScorer introduces a tree-size insensitive encoding, named \emph{epitome}. Moreover, it leverages a node merging strategy that evaluates just once nodes sharing the same threshold on the same feature. By doing so, RapidScorer outperforms  QuickScorer when dealing with a large number of leaves.

\subsection{Model Compression}
\label{subsec:modelcompr}
The effectiveness of Deep Neural Networks (DNNs) comes at the cost of a high computational complexity~\cite{nnstats}, hindering the deployment and the usage of DNNs, especially for resource-constrained devices. An inherent feature of DNNs is \textit{over-parameterization}, \textit{i.e.,} the redundancy of networks parameters: it has been proven that the same performance can be obtained with just a portion of the original parameters~\cite{denil2013predicting}. Model Compression (MC) is a recent research field investigating effective techniques for reducing the memory impact of DNNs, their inference time, and energy consumption without affecting their accuracy, exploiting over-parameterization. 
In MC techniques, we observe the presence of several lines of research: pruning~\cite{DBLP:journals/corr/HanPTD15,DBLP:journals/corr/LiKDSG16, molchanov2019pruning,DBLP:journals/corr/HanMD15,DBLP:journals/corr/GuoYC16,yu2017scalpel,vieira2017learning,he2017channel,he2018amc}, quantization~\cite{DBLP:journals/corr/LiL16,DBLP:journals/corr/ZhuHMD16, DBLP:journals/corr/RastegariORF16,hubara2017quantized,Cai_2017_CVPR,DBLP:journals/corr/ZhouYGXC17} design of efficient architectures~\cite{DBLP:journals/corr/IandolaMAHDK16,zhang2018shufflenet,howard2017mobilenets,sandler2018mobilenetv2}, knowledge distillation~\cite{bucilua2006model,ba2014deep,DBLP:journals/corr/HintonVD15}. 

Recently, pruning has shown to be extremely effective~\cite{DBLP:journals/corr/HanPTD15,DBLP:journals/corr/LiKDSG16,DBLP:journals/corr/HanMD15,DBLP:journals/corr/GuoYC16,luo2017thinet,huang2018data,yu2017scalpel,vieira2017learning,he2017channel,he2018amc}. Pruning techniques delete useless connections in a pre-trained model, producing sparse weight tensors that are lighter to store and allow for faster inference time. Performing a retraining after pruning avoids accuracy loss, even in the case of high compression factors~\cite{DBLP:journals/corr/GuoYC16}. 
The canonical classification of pruning techniques divide them into two families: 1) element-wise pruning, which sets to zero individual weights, generating sparse weight tensors and 2) structured pruning, which prunes entire groups of weights, \textit{i.e.,} columns, filters, or even entire layers. In the latter case, the resulting network's weights still belong to the dense domain. 
In this paper, we focus on element wise-pruning techniques. These methods employ heuristics to determine what are the relevant weights of the network. In particular, \textit{magnitude-based} heuristics work by removing low absolute-value weights and are proved to be effective~\cite{DBLP:journals/corr/HanPTD15,DBLP:journals/corr/GuoYC16}. In their na\"ive version, magnitude based approaches remove a fixed percentage of weights from the original model (\emph{level pruning}). Han \emph{et al.} show that the gradual increase of the target sparsity, interleaved with a number of steps of re-training, can improve the accuracy of the final model~\cite{DBLP:journals/corr/HanPTD15}. Furthermore, they propose a layer-wise threshold-based method to determine whether a parameter shall be kept or not. For each layer, its threshold $t_i$ is computed as as $t_i = \sigma_i * s_i$, with $\sigma_i$ the standard deviation of weights distribution and $s_i$ a sensitivity parameter to be chosen. By assuming that parameters follow a Normal distribution $ \mathcal{N}(0, \sigma^2) $, setting $s_i = 1$ would approximately prune away about the 68\% of the weights. The pruning step is followed by a number of re-training epochs on the surviving weights. The procedure can be then iterated by gradually increasing $s_i$ thus inducing higher sparsity. The Distiller Framework~\cite{nzmora2019distiller} version that we adopt, keeps this threshold fixed, relying on the fact that as the tensor is pruned, more elements are pulled towards the center of the distribution and then pruned. Pruning techniques have shown to be able to sparsify state-of-the-art neural architectures up to 90\%, thus strongly reducing their memory burden and easing the transmission and deployment on resource-constrained devices.

\section{Training by Scores Approximation}
\label{sec:cohen}
In this section, we detail the methodology proposed by Cohen \textit{et al.}~\cite{cohen2018universal} to train neural models approximating ensembles of regression trees.
Their technique can be considered as a special case of Knowledge Distillation~\cite{ba2014deep,DBLP:journals/corr/HintonVD15}.
Knowledge distillation is a training technique in which a small \emph{student} model is trained to mimic the outputs of a large and expressive \emph{teacher} model.
In the case of Cohen \emph{et al.}, the ensemble of regression trees plays the role of the teacher, while the neural network is the student model.
The core idea of their approach is to treat the tree-based model as a black box producing accurate scores. Formally, let us consider a Learning to Rank dataset $D = (X, Y)$,  $X \in \mathbb{R}^{f \times |D|}$, where $f$ is the number of extracted features per document, $|D|$ is the cardinality of the dataset, and $Y \in \mathbb{N}^{|D|}$ is the set of ground-truth relevances of a document w.r.t. a query.
Let $F: \mathbb{R}^{f} \rightarrow \mathbb{R}$ be the underlying function learned by an ensemble of regression trees during the training that maps a single document $x \in X$ into a relevance score.
If the neural model can reproduce the function $F$, it achieves the same ranking quality as the original model. The effectiveness of this approach relies on theoretical results showing that NNs can approximate continuous~\cite{hornik1991approximation} and piecewise continuous functions~\cite{llanas2008constructive}. In practice, the approximation is implemented by using the \emph{Mean Squared Error} as loss function computed between the network prediction and the ensemble prediction. Furthermore, the training procedure is enriched with a data augmentation step which enforces the approximation capabilities of the neural network. Consider the set of $f$ features in the dataset. For each feature, Cohen \emph{et al.}~\cite{cohen2018universal} build a list composed of 
the split points corresponding to that feature in the ensemble of regression trees, and, in the same list, they also put the maximum and the minimum for that feature in the training set.
This way they obtain a set of $f$ lists, where $f$ is the number of the features in the dataset. Each of these lists is then sorted, and replaced with its ordered midpoints, \emph{e.g.}, each adjacent pair $\{x_i, x_{i+1}\}$ is replaced with its midpoint, $\frac{x_i + x_{i+1}}{2}$.  At each training step, half of the training data is built by randomly sampling from this feature-wise set of lists to have a better coverage of the whole feature space. Before feeding them to the network, all the training data are normalized by subtracting the mean and by dividing by the variance ($Z$-normalization).
This approach is more proficient than directly learning the ground-truth relevance~\cite{cohen2018universal}. As detailed in Section \ref{sec:introduction}, the approximation error introduced is small but statistically significant in terms of ranking quality. In Section~\ref{sec:neuraleng}, we show how to mitigate this effect.

%% file: 03.methodology.tex

\section{Modeling Matrix Multiplication}
\label{sec:ModelMatMult}
In this section, we detail the optimization of matrix multiplication on modern CPUs. We start with the implementation of dense-dense matrix multiplication (DMM) and then we move to the sparse-dense (SDMM) matrix case. Matrix multiplication has a prominent role in a wide spectrum of scientific applications (linear algebra, physics, economics, engineering), and it also represents the structural operation in neural network forward and backward pass. We believe that, when dealing with the efficiency-effectiveness trade-off, a comprehensive analysis of the underlying multiplication mechanisms is essential. We develop time predictors for matrix multiplication both in the dense and in the sparse domain, and we then jointly apply them to develop an analytical model that estimates the scoring time of a neural network given the matrix shapes and the sparsity percentage of each layer of the Feed Forward Network (FFN). Our predictors are analytic, \textit{i.e.}, not learned, and they are based on 1) the knowledge gained from the implementation of DMM and SDMM on modern CPU architectures, 2) empirical measurements showing the performance of CPU on these operations under different conditions. 
 We observe that, by exploiting the predictors we are proposing, we are allowed to train only the architectures that match the desired efficiency constraints. In a latency-bound application, the efficiency constraints are specified in the requirements. In an effectiveness-oriented context, they can be inferred by observing the execution time of the competitor, \textit{i.e.}, ensembles of tree-based models. As a consequence, the use of our predictors allows to significantly reduce the search space of the optimal architecture. Furthermore, our predictors are task-agnostic, hence they can be applied in any Feed Forward Network (FFN) application field.


\subsection{Dense Matrix Multiplication}
\label{subsec:dmm}
In this section, we investigate how Dense Matrix Multiplication (DMM) is optimized on modern CPUs. DMM has countless applications, hence lots of effort has been spent to attain fast implementations. The current state-of-the-art algorithm for DMM is the the well-known Goto Algorithm~\cite{goto2008anatomy}, on which are based several open (GotoBLAS~\cite{goto2008anatomy}, OpenBLAS~\cite{xianyi2012openblas}, BLIS~\cite{huang2016blislab}) or commercial (Intel MKL~\cite{wang2014intel}) implementations.

\begin{figure}[htb]
	\centering
	\includegraphics[width=\columnwidth]{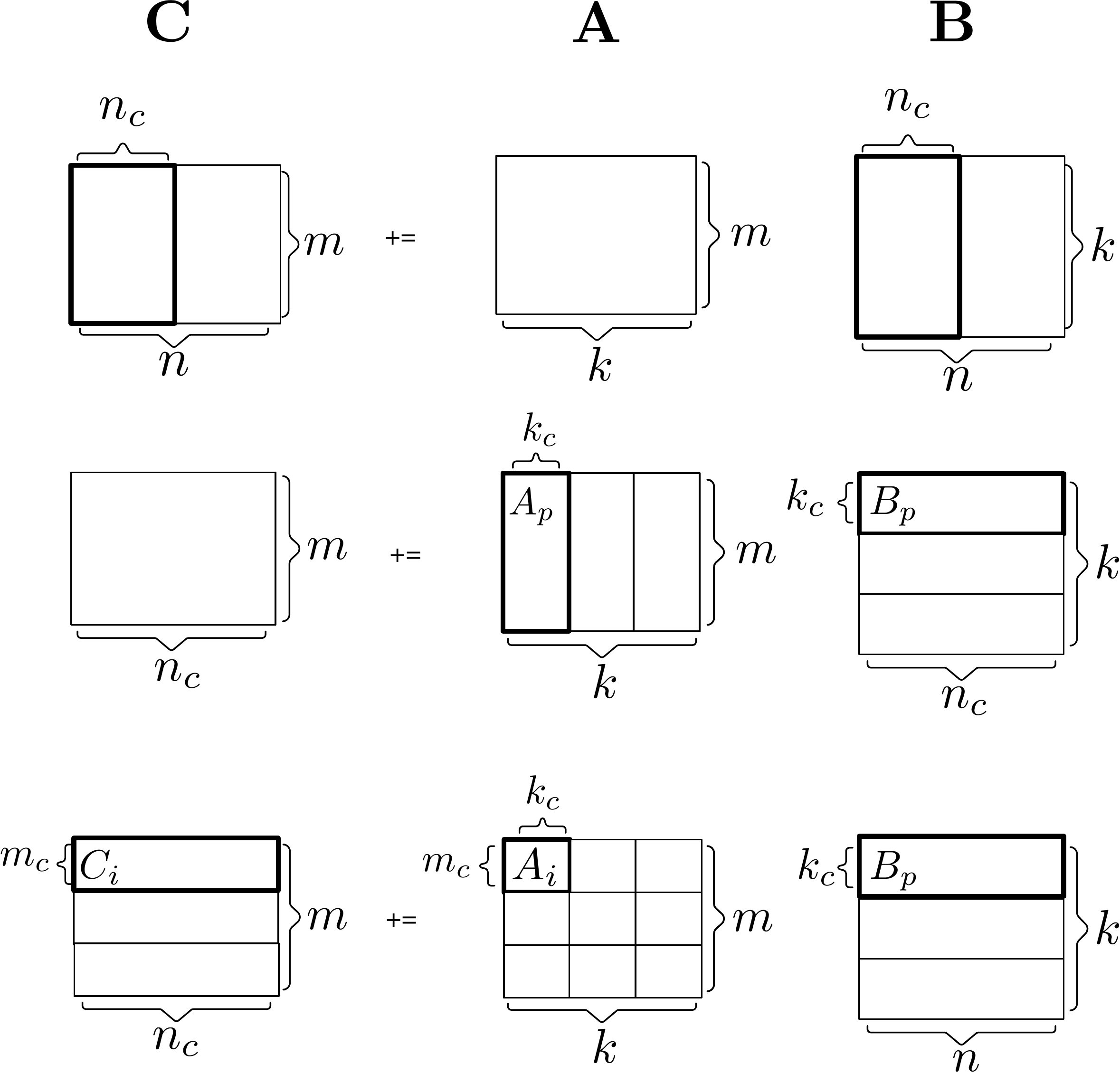}
	\caption{First three steps of the Goto algorithm for Dense Matrix multiplication.}
		\label{fig:gotofirst}
\end{figure}

The multiplication of two $ n \times n$ dense matrices involves $\mathcal{O}(n^3)$ floating-point operations with $\mathcal{O}(n^2)$ data, as can be easily evicted from Equation~\ref{eq:mmdef}. In modern processors, the interaction with memory is more time-consuming than the  computation itself (\textit{memory bandwidth bottleneck}), but a wise memory management allows to amortize the data movement over a large number of computations.
The mathematical definition of matrix multiplication is the following: given  $A \in \mathbb{R}^{m \times k}$, $B \in \mathbb{R}^{k \times n}$, the matrix multiplication binary operator computes $C = A*B$  with $C \in \mathbb{R}^{m \times n}$, where every element of $C$ is given by

\begin{equation} \label{eq:mmdef} 
C_{i,j} = \sum_{p=1}^{k} A_{i, p}  B_{p,j} \qquad  i=1, \dots, m \quad j=1, \dots, n 
\end{equation}
The Goto Algorithm consists of iteratively decomposing the overall DMM into a series of smaller matrix operations in a cache-aware fashion, until matrices fit the CPU registers. Then matrices are multiplied by means of a highly engineered \textit{micro-kernel}. We now provide a breakdown of the Goto Algorithm as implemented in the BLIS library~\cite{lawson1979basic,van2015blis}, which assumes the CPU to be equipped with 3 levels of cache and vectorized instructions. The first three steps of the blocked matrix multiplication algorithm are depicted in Figure~\ref{fig:gotofirst}.

The blocked matrix multiplication algorithm begins by partitioning along the columns of $C$ and $B$ into blocks of size $n_c$, obtaining  sub-matrices of $C$ of shape $m \times n_c$ and sub-matrices of $B$ of shape $k \times n_c$. Each $C$ sub-matrix is obtained by multiplying the complete $A$ matrix with the corresponding sub-matrix of $B$. Then, the procedure partitions the columns of $A$ and the rows of $B$ into blocks of size $k_c$, to obtain $A_p$, \textit{i.e.}, vertical panels of size $m \times k_c$, and $B_i$, \textit{i.e.}, horizontal panels of size $k_c \times n$. The $B_i$ panels are packed into the L3 cache reordering data according to a specific pattern which allows to access data contiguously even after the subsequent partitions.  We adopt the notation $\tilde{X}$ to indicate that the sub-matrix $X$ respects this pattern. Observe that, after the blocking on the $k$ axis, the original multiplication is boiled down into a series of rank-k updates so that $C = C + A_p B_p$. A further partition is performed along rows of $A$, with size $m_c$, generating $C_i$ and $A_i$. $A_i$ is, as was $B_i$ previously, packed into $\tilde{A_i}$ in the L2 cache.



\begin{figure}
	\centering
	\includegraphics[width=0.95\columnwidth ]{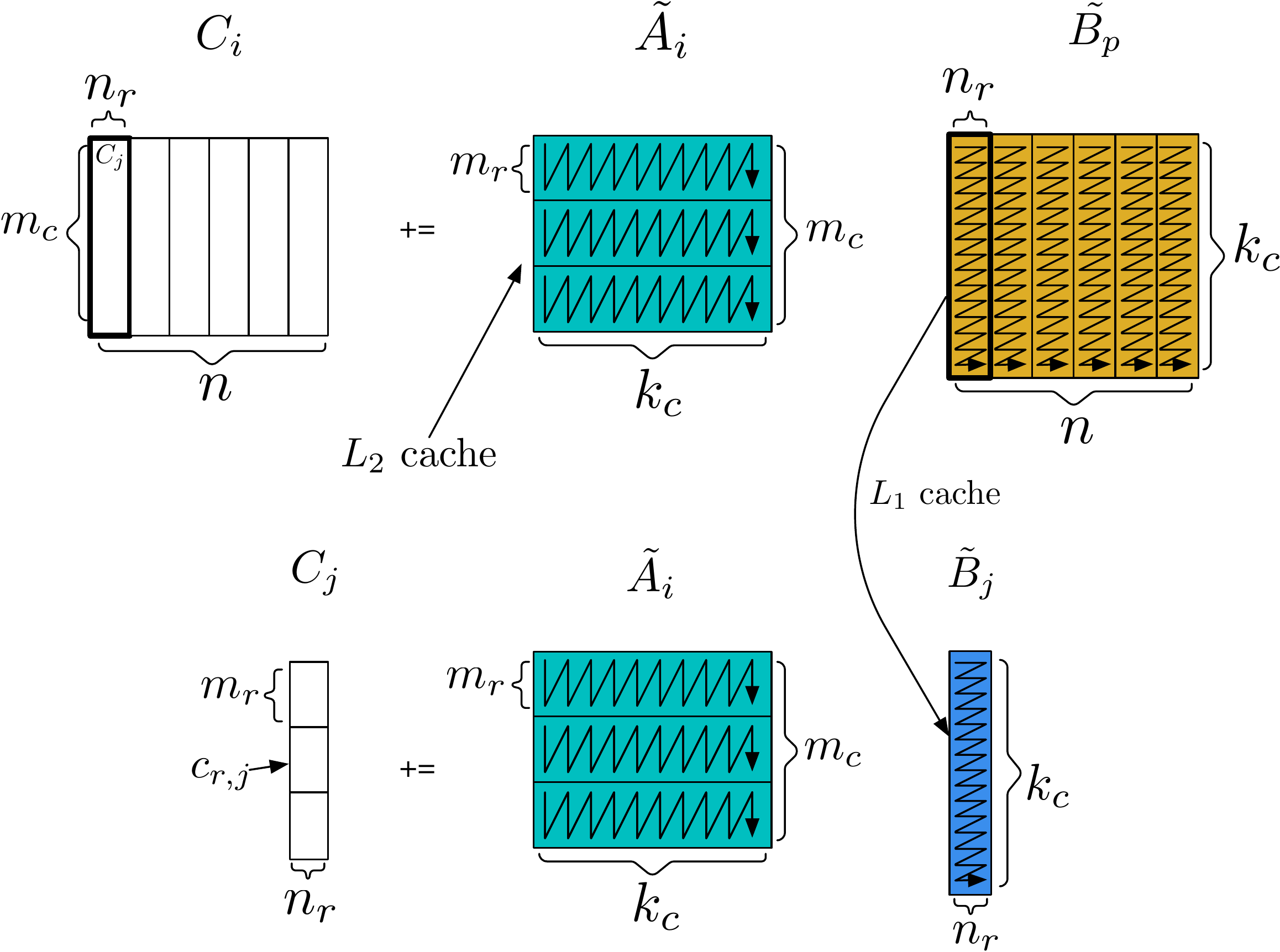}
		\caption{Macro-Kernel in the Goto algorithm for Dense Matrix Multiplication (DMM).}
		\label{fig:gotosecond}
\end{figure}

\noindent \textbf{Macro-Kernel}. The macro-kernel, or inner kernel as in the original algorithm by Goto \textit{et al.}~\cite{goto2008anatomy}, is responsible for orchestrating the memory movement between the RAM memory and the caches. Let us consider the operation $C_i \leftarrow C_i +  \tilde{A}_i*  \tilde{B}_p $, with $C_i$ of size $m_c \times n$, $\tilde{A}_i$ of size $m_c \times k_c$ and $\tilde{B}_p$ of size $k_c \times n$. The macro kernel decomposes this operation into a series of block-panel multiplications, as shown in Figure~\ref{fig:gotosecond}. As aforementioned, both $\tilde{A}_i$ and $ \tilde{B}_p$ are packed with a special pattern, indicated by the arrows in Figure~\ref{fig:gotosecond}. In particular, $\tilde{A}_i$ is organized into sub-matrices $\tilde{A}_j$ of size $m_r \times k_c$, with elements stored in column-major order, while $ \tilde{B}_p$ is organized in panels of size $k_c \times n_r$, stored in row-major order, named $\tilde{B}_j$. This data access pattern reflects the order in which the micro-kernel accesses data. 

Goto \textit{et al.}~\cite{goto2008anatomy} observed the advantages of packing $\tilde{A}_i$ into the L2 cache. The ratio between FLOPs and memory operations, regardless if the original data rely in L3 cache or in main memory, can be modeled as
 $$ \frac{2 m_{c} k_{c}}{\left(2 m_{c}+k_{c}\right)} $$
if $k_c << n$.
Hence, the higher is the $m_c k_c$ product, the smaller is the overhead of memory transfer on the overall computation. 
Knowing that L2 cache is larger than L1, we can afford larger $m_c$ and $k_c$ values \footnote{In the original work, Goto et al.~\cite{goto2008anatomy} point out that $C_i \leftarrow C_i +  \tilde{A}_i  \tilde{B}_p $ should be computed at the peak rate of CPU. This condition is true if all three matrices reside in L1 cache, but it can be considered true even if $\tilde{A}_i$ is in L2.}.

\begin{figure}[t]
	\centering
	\includegraphics[width=0.8\columnwidth ]{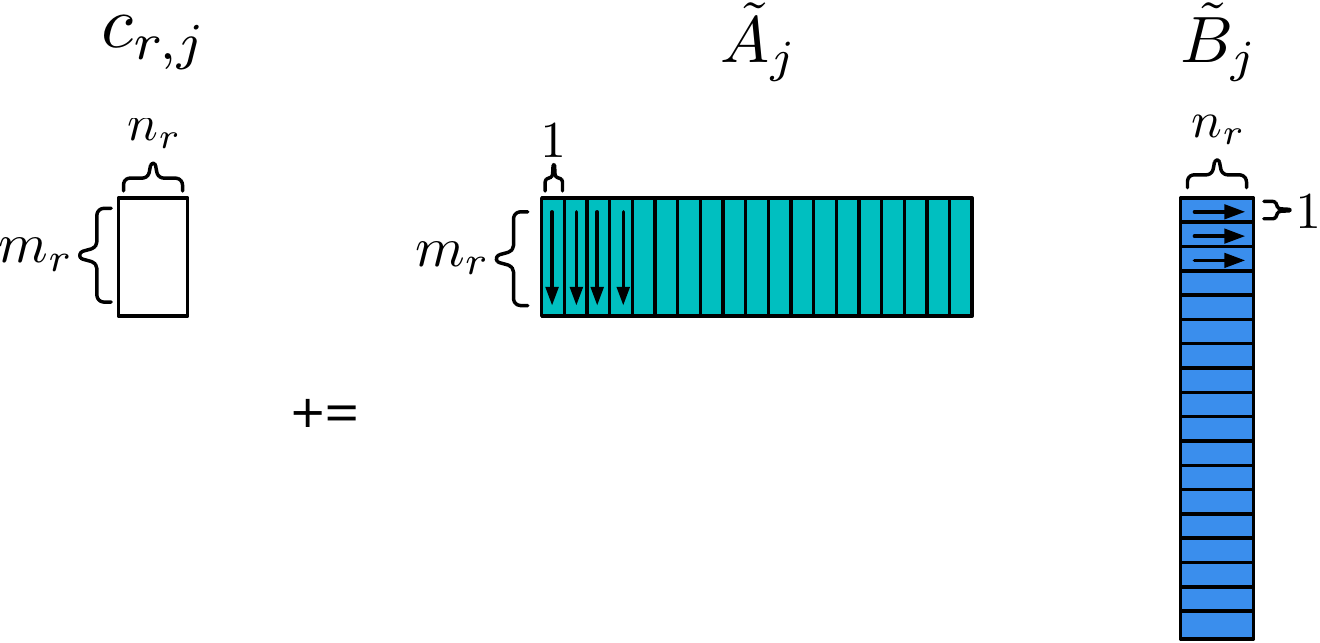}
	\caption{Micro-Kernel in the Goto algorithm for Dense Matrix Multiplication (DMM).}
	\label{fig:gotothird}
\end{figure}

\noindent \textbf{Micro-Kernel}. The micro-kernel is the core operation of blocked matrix multiplication and the speed of the whole routine largely depends on the speed of this kernel. For this reason, in high-performance libraries, the micro-kernel is often written in assembly language, to exploit vectorized instructions and hand-tuned data pre-fetching~\cite{van2015blis}.
The micro kernel computes $c_{r,j} = c_{r,j} + \tilde{A}_j \tilde{B}_j$, where $\tilde{A}_j$ is an horizontal micro-panel of $\tilde{A}_i$ and $\tilde{B}_j$ is a vertical micro-panel of $\tilde{B}_p$, residing, respectively, in L2 and L1 cache, as reported in Figure~\ref{fig:gotothird}. The operation is performed as $k_c$ rank-1 updates, by computing the outer product between a column of  $\tilde{A}_j$ and a row of $\tilde{B}_j$ and by accumulating the results into the $m_r \times n_r$ $c_{r,j}$ submatrix. In this way, $c_{r,j}$ can be kept in CPU registers until the loop over $k_c$ is , allowing to move data from the registers to the memory just once. This means that $2m_c n_c k_r$ FLOPs can be performed with just $m_r n_r$ memory operations. Furthermore, this data reading pattern benefits from the data packing performed in the previous loops. In fact, columns and rows of $\tilde{A}_j$ and $\tilde{B}_j$ respectively will be accessed contiguously, which is generally known to be faster than accessing non-in-stride memory locations~\cite{low2016analytical}. In conclusion, pre-fetching instructions that load successive entries of $\tilde{A}_j$ and $\tilde{B}_j$ are interleaved with instructions performing the rank-1 update. This allows to mask the latency of the caches with the computation time of the CPU.

\noindent \textbf{Choosing the  kernel parameters}. Blocked matrix multiplication requires to determine a number of parameters $n_c, m_c, k_c, n_r, m_r$, controlling how the matrices are gradually decomposed. These parameters can differ from one processor to another, since they are influenced by hardware features such as the cache size or the number of SIMD registers. Choosing the optimal parameters for a given CPU architecture is a research problem, tackled for example by \textit{Low et al. }~\cite{low2016analytical}, which goes beyond the scope of this paper. Here, we want to list some general rules governing good choices for parameters. 
The micro-kernel is characterized by $m_r, n_r, k_c$. The values of $m_r$ and $n_r$ should be large enough so that the computation masks the latency of the caches. However, it should also allow to leave space in the registers for the next entries of $\tilde{A}_j$ and $\tilde{B}_j$. $k_c$ should be as large as possible, but must take into account the following constraints: 1) $k_c n_r$ entries from $\tilde{B}_j$ should fit the L1 cache 2) $m_c k_c$ entries from $\tilde{A}_i$ reside in the L2 cache. 
Moreover, cache replacement policies should also be taken into account. These policies control which data are kept and which are discarded from the levels of cache and may impact on the optimal macro-kernel values. A general solution is provided by Goto \textit{et al.}, who suggest choosing $k_c$ so that $\tilde{B}_j$ takes less than the half of the L1 cache~\cite{goto2008anatomy}.

%
Concerning the macro-kernel, we already discussed that the $m_c k_c$ product should be as large as possible. One of the key insights of the Goto algorithm is to consider the role of the Translation Look-Aside Buffer (TLB) in choosing the macro-kernel parameters. To hide the limits of random-access memories capacity (RAM), modern computing architectures use virtual memory. With this mechanism, the memory (RAM and hard disk) is partitioned into pages and a table, called \textit{page table}, keeps track whether a page is in memory or on disk. Scanning the page table entails additional overhead to check whether the requested page is on memory or disk.
Hence, the TLB, which is smaller than the overall \textit{page table}, keeps track of the most recently used pages: in case of a TLB \textit{hit}, the translation is fast. On the other side, in case of a TLB \emph{miss}, the complete \textit{page table} is checked and the new entry is moved to the TLB. Actually, the TLB has the same role as the cache and the \textit{hit/miss} dichotomy involves the same consequences. Thus, besides ensuring that $m_c k_c$ entries from $\tilde{A}_i$ fit the L2 cache, it is crucial that $\tilde{A}_i$, $n_r$ columns from $C_k$, and $n_r$ columns of $\tilde{B}_j$ are simultaneously addressable by the TLB, to avoid TLB misses during the block-panel multiplications of the macro-kernel. The only limit to the $n_c$ parameter is that $k_c n_c$ have to fit the L3 cache.

\subsection{Dense Neural Forward Pass Time Predictor}
\label{subsec:densetimepred}
In the previous section, we detail how Dense Matrix Multiplication is implemented on modern CPU architectures. We now show how the insights deriving from a deep understanding of matrix multiplication can be used to develop a time predictor for a Feed Forward Network (FFN) forward pass. We empirically demonstrate that even the highly engineered Goto algorithm  suffers when dealing with edge matrix dimensions. Hence, we leverage this intuition to build a hybrid analytical-empirical model for predicting dense matrix multiplication.
A FFN is composed of a  stack of \textit{fully connected} layers, where each neuron of layer $i$ is connected to all neurons of layer $i+1$. Each layer is composed of a weight matrix $W_i$, a bias vector $b_i$ and a non-linear \textit{activation function} $\sigma_i( \cdot )$. Let $x_i$ be the input to the $i$-th layer, the forward pass of layer $i$ is described by:
\begin{equation}
	\label{eq:mlpforward}
	x_{i+1} = \sigma_i(W^t_i x_i + b_i)
\end{equation}
where $x_{i+1}$ represents the output of the $i$-th layer. 
Hence, forwarding through a FFN layer consists of: 1) multiplying the input with the weight matrix, 2) summing the bias, 3) applying a non-linear activation function, usually ReLU or its variants. The overall forward pass on a FFN of $d$ layers has a cost, in terms of execution time, given by:
\begin{align}
 \label{eq:overallcost}
 	T = t_m \cdot ( f \cdot l_1 + \sum_{i=2}^{d} l_i   l_{i-1} + l_{d})
 	 + t_a \cdot \sum_{i=1}^{d} l_i + t_r \cdot \sum_{i=1}^{d} l_i \nonumber \\
 	   \simeq t_m \cdot ( f \cdot l_1 + \sum_{i=2}^{d} l_i \cdot  l_{i-1} + l_d) 
 \end{align}
where $t_m $ is the normalized time per multiplication, $t_a$ is the time for addition, $t_r$ is the time to perform the ReLU operation on a single neuron. As reported in Equation~\ref{eq:overallcost}, the time to perform matrix multiplication dominates the overall execution time, both in terms of number of operations and in terms of the complexity of the operation itself. We observe that $t_m$ can be inferred as:

\begin{equation}
	\label{eq:tm}
	t_m = \frac{1}{\text{GFLOPS}}
\end{equation}
The theoretical peak of GLOPs can be derived form the hardware specifications of the processor\footnote{https://software.intel.com/en-us/articles/a-simple-example-to-measure-the-performance-of-an-intel-mkl-function}. However, real performance can be significantly different from the theoretical ones, especially when facing limit cases, such as narrow or wide matrices. To include these cases into our evaluation, we develop a prediction model to measure the performance of a specific neural networks architecture.

Among the different instantiations of the BLAS library, we choose \textit{oneDNN}\footnote{\url{https://github.com/oneapi-src/oneDNN}},
a C++ high-performance framework for deep learning primitives developed by Intel, used as backbone inference system by Pytorch~\cite{NEURIPS2019_9015}, Tensorflow~\cite{abadi2016tensorflow}. With respect to the Math Kernel Library (MKL)~\cite{wang2014intel} by Intel, oneDNN guarantees the same performances while being open source.
The oneDNN library adopts the following parameters for CPUs with AVX2 ISA enabled: $m_c = 10000$, $n_c =384 $, $k_c = 192$, while for the micro-kernel we have  $m_r = 24, n_r = 4$.
The macro-kernel parameters $m_c, n_c, k_c$  are selected to deal with very large matrices; for the sequential case, the library contains a mechanism to tailor smaller shapes. Let us call $\overline{m}_c$, $\overline{n}_c$, $\overline{k}_c$ the parameters that the macro and micro kernels actually use. 	
$\overline{m}_c$ is chosen as:
$$\overline{m}_c = \texttt{rnd\_up}(min(max(m, m_r), m_c
), m_r)$$
where \texttt{rnd\_up}($a,b$) is a function which approximates $a$ as $a = n^{*} b$, with $n^{*} = min\{n \ |\  nb \geq a\}$, \textit{i.e.}, to the subsequent multiple of $b$. This way, it is ensured that $\overline{m}_c$ is larger than the micro-kernel parameter $m_r$ and that the default $m_c$ is not involved if $m \leq m_c$. By means of the \texttt{rnd\_up} function, we ensure that $\overline{m}_c \bmod m_r  = 0$ to avoid undersized horizontal $\tilde{A}_j$ panels in the micro-kernel.  
Similar refinements are adopted to chose $\overline{n}_c$ and $\overline{k}_c$. Moreover, oneDNN triggers when the cost of packing the matrices into contiguous arrays surpasses the cost of multiplication. In this case, besides changing the macro-kernel parameters, it also performs a different routine that skips copying the matrices in cache-aware buffers.





\begin{figure}	
	\centering
	\includegraphics[width=\columnwidth]{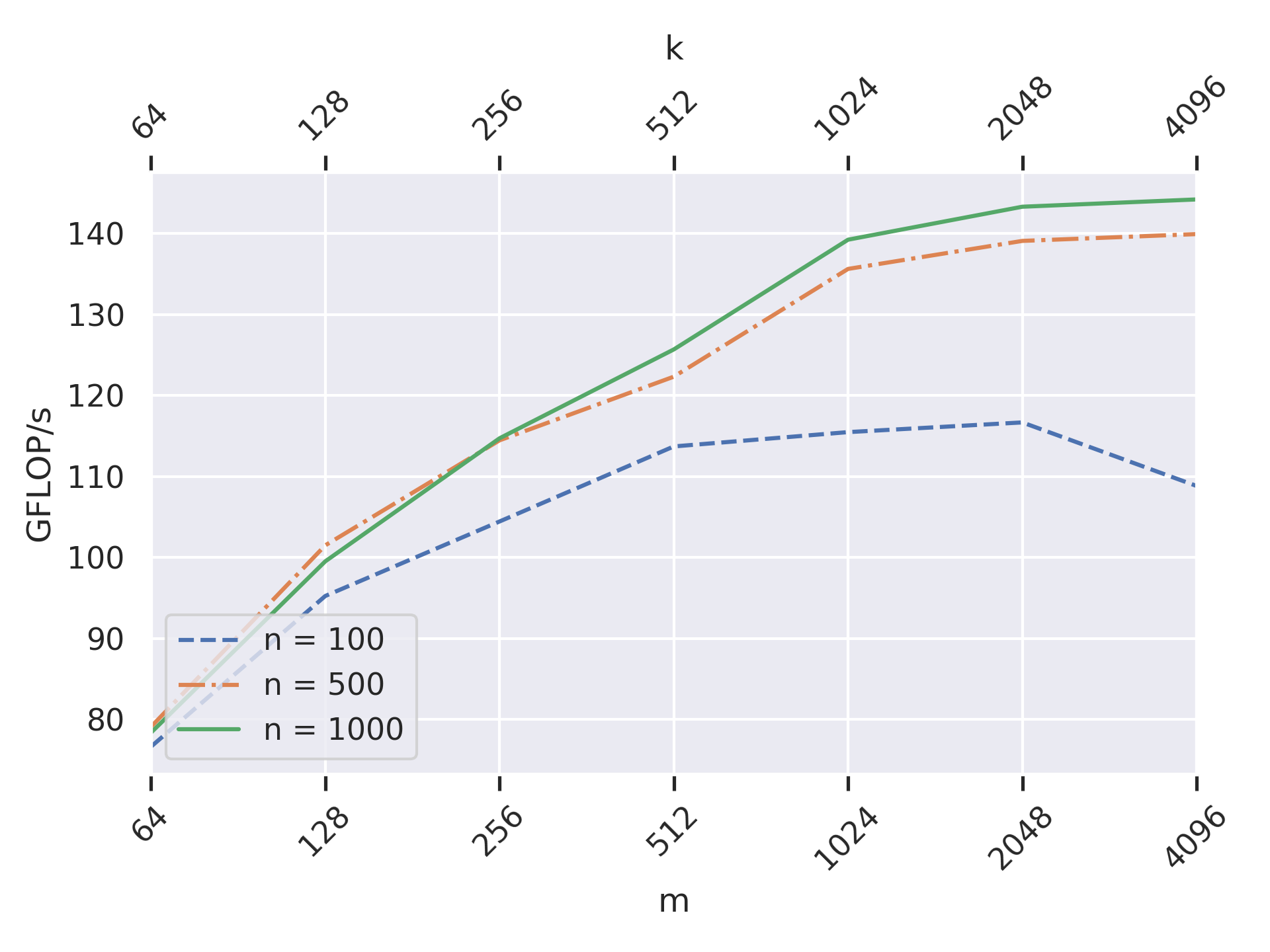}
	\caption{Matrix Multiplication with oneDNN, as $m$ and $k$ grow.  }
	\label{fig:onednnl_no_r}
\end{figure}

\begin{figure}
	\centering
	\includegraphics[width=0.8\columnwidth]{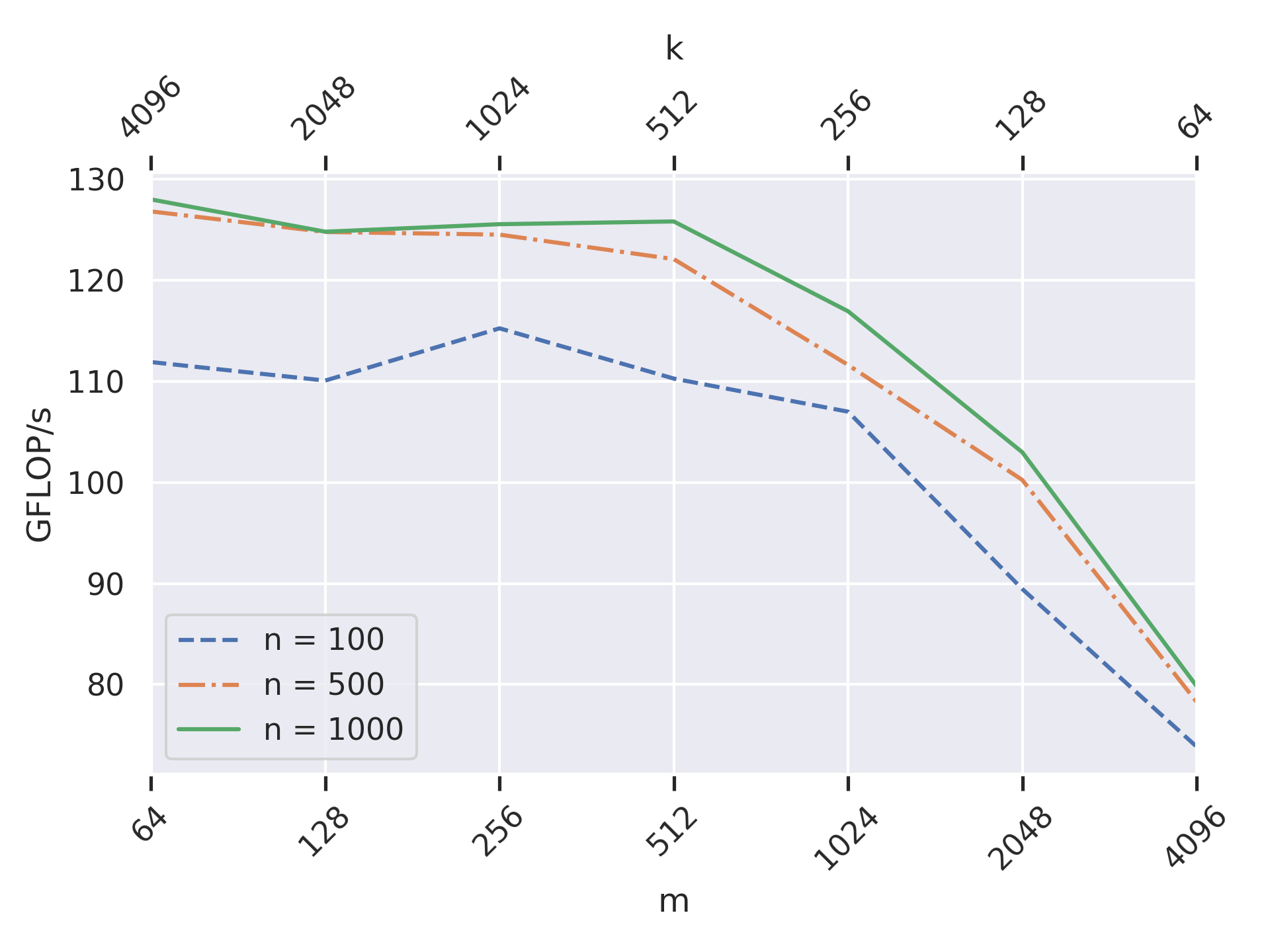}
	\caption{Matrix Multiplication with oneDNN, with the product $mk$ constant. }
	
	\label{fig:onednnl_rev}
\end{figure}

\begin{figure}
	\centering
	\includegraphics[width=\columnwidth]{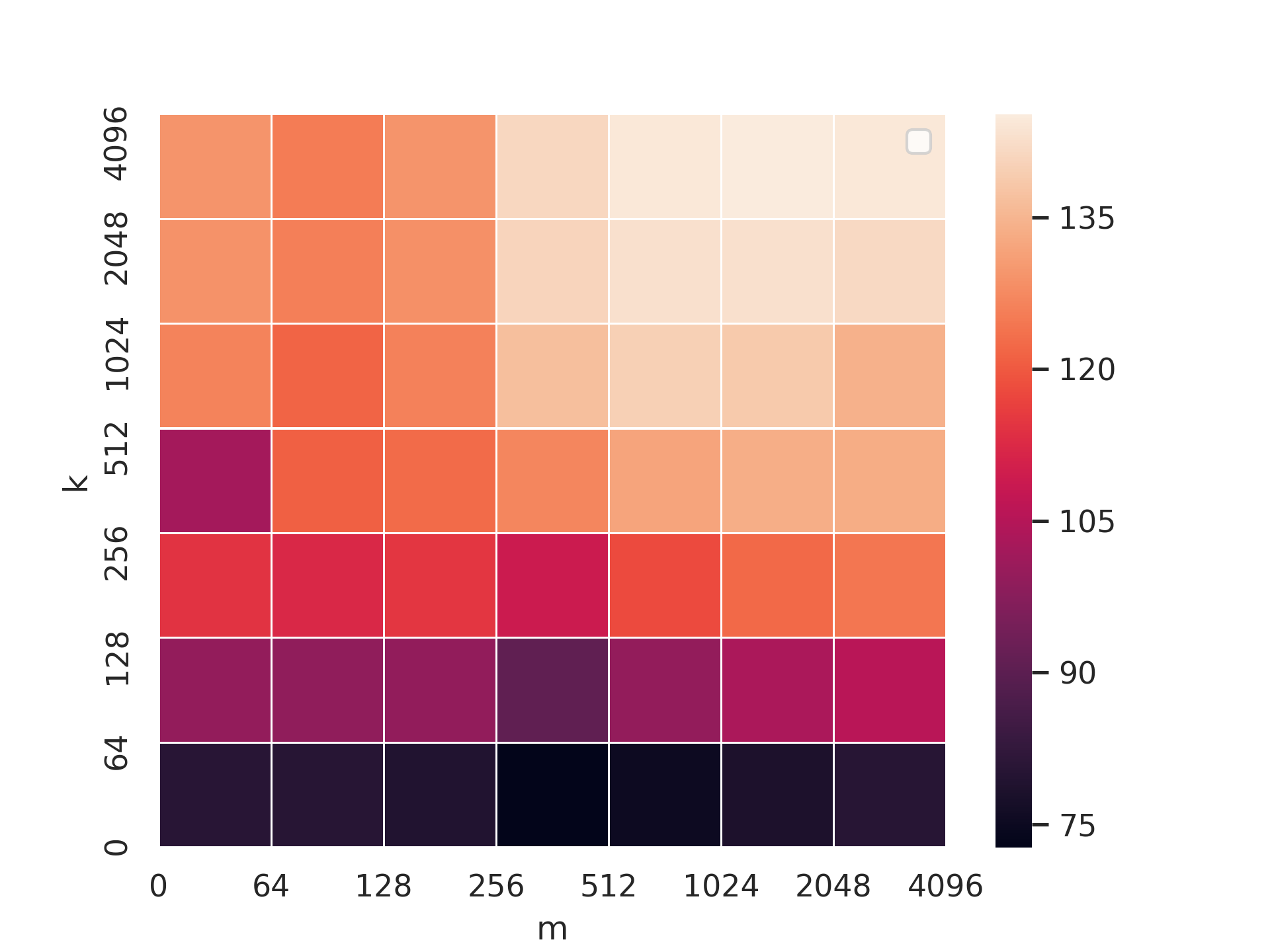}
	\caption{Matrix Multiplication HeatMap with n = 1000.}
	\label{fig:heatmap}
\end{figure}

In Section~\ref{subsec:dmm} we have described the optimization techniques beyond Dense Matrix Multiplication on modern CPUs. We also detail the tailored refinements implemented by the oneDNN framework to deal with matrices where at least one dimension is small. We now show the performance of the oneDNN framework with differently shaped matrices, aiming at identifying a reliable $t_m$ for Equation~\ref{eq:tm}. In these experiments, we multiply random matrices with different shapes to empirically analyze how the oneDNN library adapts to different matrix dimensions. We propose two different cases: 1) $m=k$, 2) $mk=c$, with $c$ as a constant integer.
We run our tests on a i9-9900K processor, with AVX2 instructions, 3.6 GHz, max frequency 5.0 GHz. Each core has a 32 KiB L1 cache for data, 32 KiB L2 cache for instructions, both 8-way set associative, 256 KiB L2 cache 4-way set associative, and 2 MiB L3 cache, 16-way set associative. We report the results for single-thread execution. 
In our first experiment, we vary $m$ and $k$ in a fixed range and report the corresponding GFLOPs, with different values of $n$. Observe that $A$, of shape $m \times k$,  represents the weight matrix $W$, $B$, of shape $k \times n$ represents the input matrix $x$, obtained by stacking $n$ input vector. We vary $m$, $k$ and $n$ to model real use-case scenarios: $m$ and $k$ correspond to the sizes of Feed Forward Network layers, while $n$, the \emph{batch size}, is the number of documents we give in input to the neural network at a time. Results are reported in Figure~\ref{fig:onednnl_no_r}, which shows that GFLOPS grow as the size of the matrices even with the aforementioned techniques tailored to edge cases. In Figure~\ref{fig:onednnl_rev}, we show the results of the reverse experiment: instead of gradually increasing both $m$ and $k$, we keep the size of $A$ constant (the $mk$ product is constant). The figure shows that small values of $m$ with large values of $k$ still afford  high-performance (left side of the graph). On the other hand, small values of $k$ paired with larger values of $m$ cause serious performance degradation. The variation of the GFLOPS with the matrix shapes suggests that a unique and size-independent $t_m$ is not reliable. As aforementioned, this evidence some limitations of the Goto algorithm when dealing with edge combinations of input dimensions.
A correct analysis expresses $t_m$ as a function of the $m,n,k$ parameters, or in the case of the Feed Forward Network , as a function of the dimensions of the layers, \textit{i.e.}, $t_m = t_m( l_1, \dots, l_{d})$.
Given the variability of the performance with input shapes, we shall empirically measure them.
 We can use Figure~\ref{fig:onednnl_no_r} and ~\ref{fig:onednnl_rev} to derive a lookup table that maps the matrix shapes to the corresponding GFLOPs. The previous graphs are synthesized in Figure~\ref{fig:heatmap}, which shows an heatmap of the GLFOPs with different values of $m$ and $k$ and $n = 1000$.

 We observe three performance zones, defined by horizontal stripes induced by partitioning the $k$ axis.
\begin{itemize}
	\item $K \geq 512$ : high-performance (130 GFLOPs)
	\item $128 \leq K \leq 512$: Medium performance (110 GFLOPS)
	\item $K \leq 128$: Low performance (90 GFLOPs)
\end{itemize}


For a network of size $\{1000, 500, 500, 100 \}$, we can assume to be always in the high-performance region, except for the last layer. Observe that the last layer has a negligible impact on the overall forward time and we can ignore it. Table~\ref{table:est_vs_real_exec_t} illustrates how the prediction model can substitute the experimental procedure of training and testing a model, turning out to be essential to reduce the architecture search space.

\begin{table}[htb]
	\centering

	\begin{tabular}{lrr}
		\toprule
		\multirow{2}{*}{Model} &    \multicolumn{2}{c}{Scoring Time ($\mu s$/doc)} \\
		\cmidrule{2-3}
		 & Real & Predicted \\		
		\midrule

		 1000$\times$500$\times$500$\times$100 & 14.4& 14.5 \\
		 200$\times$100$\times$100$\times$50 & 	1.3& 1.3 \\
		 300$\times$150$\times$150$\times$30 & 2.0 & 2.2 \\
		 500$\times$100 & 2.1 & 2.2\\
 		\bottomrule
	\end{tabular}
	\caption{Performance of our dense prediction model. Real execution times measured with batch size = 1000.}
	\label{table:est_vs_real_exec_t}
\end{table}

\subsection{Sparse-Dense Matrix Multiplication} 
\label{subsec:sdmm}
In this section, we study Sparse-Dense Matrix Multiplication (SDMM), a special case of matrix multiplication where the first matrix is \textit{sparse}: we recall that \textit{sparsity} is defined as the percentage of zero entries in a data structure, in this case, a matrix. First, we describe a common format to store sparse matrix, \textit{Compressed Sparse Row} (CR). Then, we detail how SDMM is implemented on modern CPU processors. 

\smallskip
\noindent \textbf{CSR Format}.
A sparse matrix is completely identified by its non-zero values and their positions since all the others entries are zeros. This motivates the use of a different representation for sparse matrices w.r.t. to dense ones. The different representation aims at saving storage space and improving the performance of matrix multiplication. For this purpose, several formats have been developed: the most common are Compressed Sparse Row (CSR), Compressed Sparse Column (CSC), Coordinate List (COO). Among them, we analyze CSR, since it is usually supported by off-the-shelf libraries, both for storing and for matrix operators, such as multiplication and it naturally fits to Sparse-Dense Matrix Multiplication, as we will detail.

\begin{figure}
\centering
	\includegraphics[width=0.8\columnwidth]{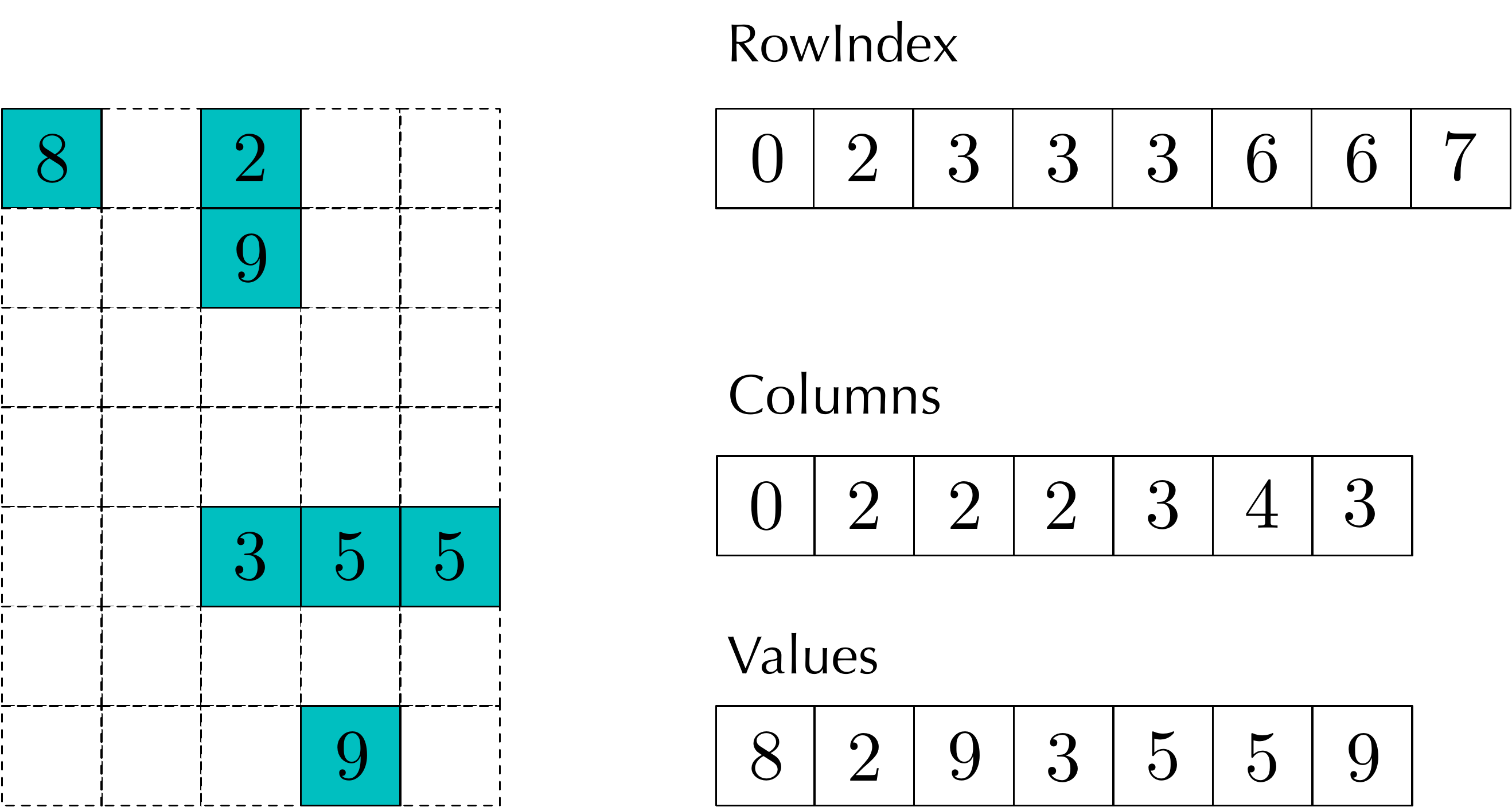}
	\caption{CSR Format for Sparse Matrices.}
	\label{fig:sparsecsr}
\end{figure}

Let us consider a matrix $M \in \mathbb{R}^{m \times n}$ with $nnz$ non-zero elements. 
An example of the  CSR representation is reported in Figure~\ref{fig:sparsecsr}. It consists of three vectors: \textit{values} $\in \mathbb{R}^{nnz}$, \textit{columnIndex} $ \in \mathbb{R}^{nnz}$, \textit{rows} $\in \mathbb{R}^{m+1}$. 
The \textit{values} array stores the non-zero entries, and \textit{columnIndex} stores their column index in the original matrix, meaning that \textit{columnIndex}$[i]$ stores the columns index of \textit{values}$[i]$. The \textit{rows} array is built so that $rows[i+1] - rows[i] $ is the number of non-zero entries for row $i$. 

\smallskip
\noindent \textbf{Sparse Dense Matrix Multiplication}.
Sparse Dense Matrix Multiplication or sparse Multi-vector multiplication (SDMM) has a large range of applications: fluid dynamics, graph analysis~\cite{tiskin2001all}, non-negative matrix factorization~\cite{kim2011fast}, economic modeling, seismic simulations~\cite{breuer2019petaflop}, and machine learning~\cite{NIPS2010_4099}. Pruning a neural network pre-trained model naturally induces the usage of SDMM in the forward pass of a Multi-Layer Perceptron, since it converts dense weights into sparse ones. Let us consider Equation~\ref{eq:mlpforward}: in the most general case $W$ represents the dense weight matrix. After pruning, $W$ is transformed into a sparse matrix $\dot{W}$, thus converting $\dot{W}^T x$  into a Sparse Dense Matrix Multiplication.

Consider the operation $C = AB$, where $A \in \mathbb{R}^{m \times k}$ is a sparse matrix in the CSR representation with $nnz$ non-zero values, and $B \in \mathbb{R}^{k \times n}$, $C \in \mathbb{R}^{m \times n}$ are dense matrices. 
The mundane algorithm induced by A being in CSR Format is reported in Algorithm~\ref{alg:csr_mult}. This format is suitable for row-wise access, allowing to consider exclusively the non-zero entries of the left-side matrix. 
The total number of floating-point operations is reduced from $2mnk$ to $2 nnz N$ w.r.t. dense case, but the irregular access pattern induced by sparsity hinders the efficiency of the algorithm. To overcome this problem, a twofold strategy, as for the dense case, is applied: 1) proficient data access pattern, 2) optimization of the core operation (\textit{micro-kernel}). 

The most used library for sparse matrix multiplication is the Math Kernel Library (MKL)~\cite{wang2014intel}, which implements the sparse versions of third level BLAS routines. Since the library is closed and there are no details on how the multiplication is implemented\footnote{https://community.intel.com/t5/Intel-oneAPI-Math-Kernel-Library/Sparse-Dense-Matrix-Multiplication/m-p/1173953}, we refer to the implementation of the LIBXSMM~\cite{heinecke2016libxsmm}, which is open-source. Later on in this section, we show that LIBXSMM actually outperforms MKL in the spectrum of shapes involved by our neural networks. 

\begin{figure}
\centering
	\includegraphics[width=\columnwidth]{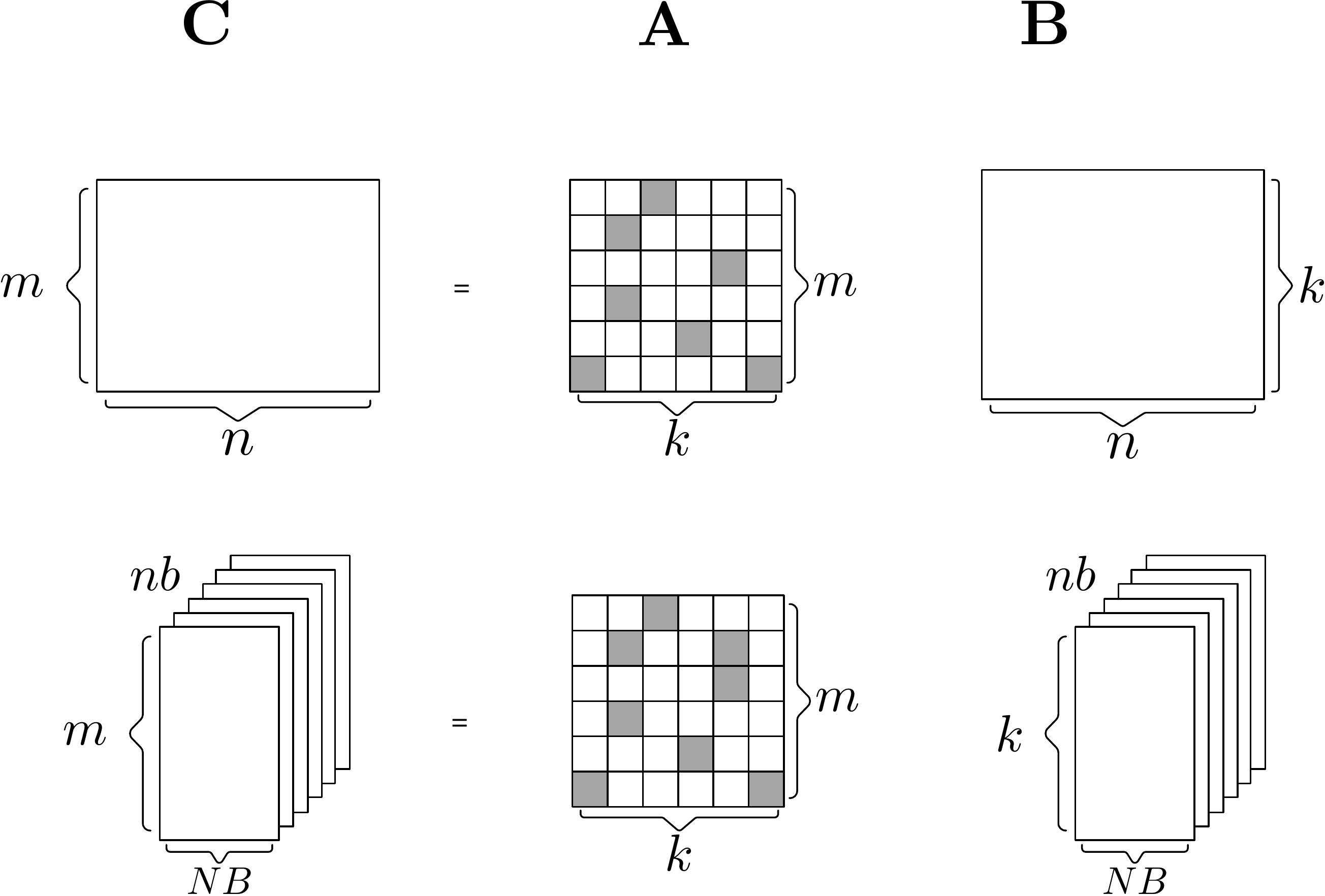}
	\caption{LIBXSMM Sparse-Dense Matrix Multiplication (SPMM).}
	\label{fig:libxsmmsparsedense}
\end{figure}

\begin{algorithm}[]
	\KwData{ CSR $A \in \mathbb{R}^{m \times k}$, $B \in \mathbb{R}^{k \times n}$}
	\KwResult{$C \in \mathbb{R}^{m \times n}$  }
	$C[i,k] = 0$\;
	\For{$i = 0$ \KwTo $M-1$}{
		\For{ $j = A.rows[i]$ \KwTo $A.rows[i+1]-1$}{
		  \For{$k = 0$ \KwTo N-1}{
		  	$idx   = A.cols[j]$\;
		  	$C[i,k] \leftarrow C[i,k] +  A.val[j] * I [idx, k]$\;
		  }
		}
	}

	\caption{Sparse-Dense Matrix Multiplication algorithm with CSR format.}
	\label{alg:csr_mult}
\end{algorithm}


\smallskip
\noindent \textbf{Sparse-Dense matrix multiplication with LIBXSMM}. LIBXSMM~\cite{heinecke2016libxsmm} is a high-performance library specifically tailored for Intel architectures, specialized in small dense matrix multiplication, sparse matrix multiplication, and deep learning primitives in general. It is based on ``Just in Time'' (JIT) code specialization, which intends to exploit the runtime information about its operands. The sparse-dense routine was originally developed to solve seismic equations~\cite{breuer2019petaflop}.

We now detail the sparse-dense matrix multiplication as implemented in the LIBXSMM library, with A in CSR format. 
The dense matrix $B$ is converted into a three dimensional tensor of shape $k \times N_b \times n_b$, as reported in Figure~\ref{fig:libxsmmsparsedense},
so that $N = N_b \times  n_b$. This means to factorize the $N$ dimension in two sub-dimension, in which one ($n_b$) is induced by the underlying hardware. The ideal value of $n_b$ in fact, coincides with the SIMD length of the processor, \textit{i.e.}, the number of different numbers that a SIMD vector can store. 
Using floating-point variables (32 bit) on a machine with AVX2 ISA (256 bit), the SIMD length is 8. This packing allows to  multiply each non-zero element of $A$  with $nb$ values of $B$ at time, using just one vectorized instruction.

\begin{figure}[t]
\centering
	\includegraphics[width=\columnwidth]{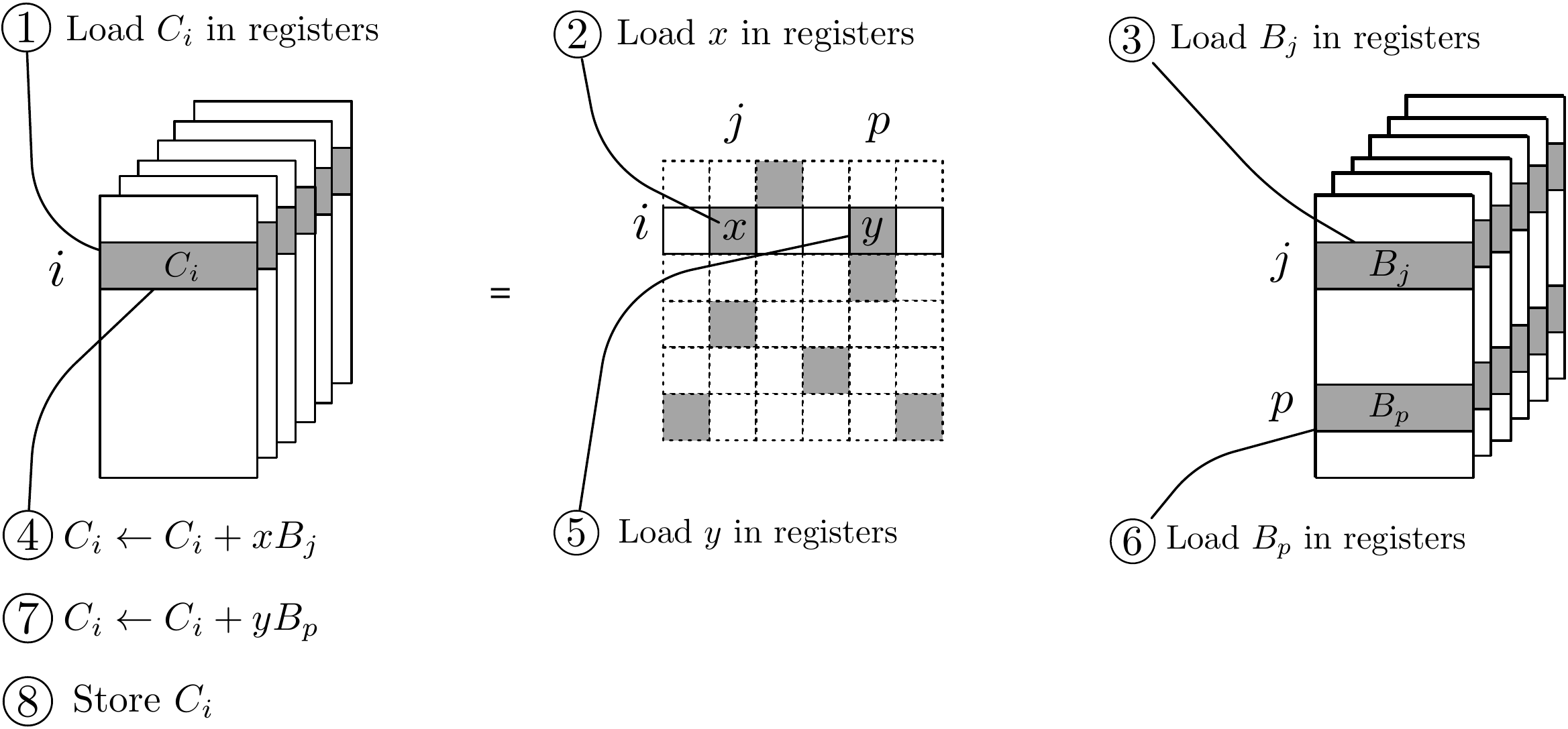}
	\caption{Micro Kernel of LIBXSMM Sparse-Dense Matrix Multiplication (SPMM).}
	\label{fig:libxsmmsparsedensemicro}
\end{figure}

The problem of irregular accesses is tackled by hardwiring the loading of the elements of $A$ and $B$, so that only relevant elements are loaded. The data access pattern provides for multiplying each non-zero element of $A$ ($a_{i,j}$) with the $j$-th rows of $B$ ($B_j$) and accumulate the results into the $i$-th row $C$ ($C_i$).
The computation is carried on one row of $A$ at time. Figure~\ref{fig:libxsmmsparsedensemicro} shows the sub-routine performed for each row $i$.  
Let us call the first non-zero element of the current row $x$, in position $(i,j)$; we assume to have at least one non-zero entry, otherwise the row is skipped. $C_i$ is loaded into $N_b$ SIMD registers, each containing $n_b$ values. $x$ is \emph{broadcasted} to a SIMD CPU register, \textit{i.e.}, $n_b$ consecutive copies of the $x$ vector are loaded into the register. We refer to this vector as $\overline{x}$. 
$B_j$ is loaded as well and $C_i$ is updated as $C_i \leftarrow C_i + \overline{x} B_j$; the update involves $N_b$ Fused Multiply Add (FMA) instructions $C_{i,k} \leftarrow C_{i,k} + \overline{x} B_{j,k}$. Then, the routine moves to the next non-zero elements in the $i$-th row of $A$. Once all the non-zero elements have been multiplied, $C_i$ is stored in memory and the algorithm moves on to the next row of $A$.

LIBXSMM is equipped with a mechanism that interrupts the code generation if the number of instructions is too elevated. This can happen if the number of non-zero elements in $A$ or the $N$ dimension are too large. Since the $N$ dimension corresponds to the batch size in the neural forward, we are free to  reduce it to overcome this limit. When necessary, we also split the $m \times k $ $A$ matrix along the $M$ dimension to generate a set of sub-matrices $A_S = \{A_1, \dots, A_s \ | \  A_i \text{ of size }  M/s \times k  \}$. Each $A_i$ will have fewer non-zero entries, preventing code generation failure. 
The $C$ matrix is trivially obtained by multiplying each $A_i \in A_S$ with $B$ separately and by stacking the results along the $M$ (vertical) axis:
\[ C = 
  \begin{bmatrix}
    \begin{array}{c}
  A_1 B  \\
  \hline
  A_2 B\\
  \hline 
  \vdots\\
  \hline
  A_s B\\	   
    \end{array}
  \end{bmatrix}
	\]

\smallskip
\noindent \textbf{LIBXSMM vs MKL}. As aforementioned, MKL is known to provide the fastest routine for sparse-dense matrix multiplication. We now show that LIBXSMM outperforms MKL on small, very sparse, and asymmetric matrices, which is the typology of matrices we employ in our MLPs for document scoring. In Table~\ref{table:lib_vs_mkl_msn}, we report the execution time of $C = AB$, with $A$ sparse in the CSR format and B dense, both for MKL and LIBXSMM; on the $x$-axis is reported the shape  ($m \times k $) $A$ and its sparsity. $B$ has shape $k \times n$, where $n$ is the batch size, set to $64$. The matrices correspond to the first layer of real models trained on the \msn dataset~\cite{DBLP:journals/corr/QinL13}, which provides $136$ handcrafted features. The Table shows that LIBXSMM is always faster than MKL on these shapes, with a speedup factor often larger than $2$x. This consideration, together with the availability of the code, has led us to pick the LIBXSMM library as the reference implementation. 

\begin{table}[htb]
	\centering
	\begin{tabular}{llrr}
		\toprule
		\multirow{2}{*}{Shape} &   	\multirow{2}{*}{Sparsity}&  \multicolumn{2}{c}{SDMM Time ($\mu s$)}\\
		\cmidrule{3-4}
		 & & MKL & LIBXSMM \\
		\midrule
		400$\times$136 &  0.996 &         3.1 &          \textbf{1.2} \\
 		300$\times$136 &  0.985 &         2.5 &          \textbf{1.4} \\
 		200$\times$136 &  0.971 &         2.8 &          \textbf{1.6} \\
 		100$\times$136 &  0.989 &         1.0 &          \textbf{0.4} \\
 		50$\times$136  &  0.968 &         0.7 &          \textbf{0.2} \\
 		\bottomrule
	\end{tabular}
	\caption{ Comparison between MKL and LIBXSMM for Sparse Dense Matrix Multiplication (SDMM)	. Shapes and sparsities represent the first layer of FFNs trained on \msn. Batch size is set to $64$.}
	\label{table:lib_vs_mkl_msn}
\end{table}


\subsection{Sparse Time Predictor}
\label{subsec:sptimepred}
In this section, we illustrate the development of a Sparse-Dense matrix multiplication time predictor, specularly for what we have done for the dense case.
As detailed in Section~\ref{subsec:sdmm}, the algorithm provides for iterating over the rows of $A$ with at least one non-zero entry. We start by analyzing the time cost of multiplying the $i$-th row of $A$ with $B$, which is given by the sum of the cost of the following operations. 

\begin{enumerate}
	\item Loading $N_b$ vectorized elements from $C$ (each one of size $n_b$).
	\item Loading each non-zero element in $A_i$. Since the non-zero values of $A$ are stored contiguously in $A.values$, this operation benefits from cache memory.
	\item Loading $N_b$ vectorized elements of $B$ (of size $n_b$) for each non zero element of $A$.
	\item Updating $C_j \leftarrow C_j + x *B_j$, for each $x\neq0$ in $A_i$. Each update consists in $N_b$ FMA instructions. 
	\item Storing $N_b$ vectorized elements of $C$ (each one of size $n_b$).

\end{enumerate}
Let us define $a_c$ as the set of active columns in $A$, namely the set of columns containing at least one non-zero element, and $a_r$ the set of active rows in $A$. Let us also define $L_c$ as the cost to load and store $N_b$ elements of $C$, $L_a$ the cost of loading one element of $A$ and updating $C_j$ with $N_b$ FMA instructions, and $L_b$ the cost of loading $N_b$ elements of $B$. 
When generalizing the previous costs to the entire matrices, we have to take into account the effects of caching. 
While $A$ and $C$ are loaded just once, $B$ elements can be loaded multiple times; whether they benefit or not from the caching mechanism depends on the access pattern induced by the non-zero entries of $A$.
For example, if $x$ in position $(i,j)$ is a non-zero element, $B_j$ needs to be loaded into the registers from the main memory and the cache will also retain a copy of $B_j$.
Assume that in a successive row of $A$ exists an element $x' \neq 0$ on the same column of $x$, \textit{i.e.} in position $(g, j)$, with arbitrary $g$. When performing $C_g \leftarrow C_g + x' B_j$,  $B_j$ already resides in the cache: since loading elements from the cache is way much faster than loading them from memory, the cost of re-loading $B_j$ can be considered negligible. 
Assuming that once a row of $B$ is loaded into the cache it remains there until the end of the operation, we pay the cost of loading a row $B_j$ just the first time that this row is loaded. At the same time, if there are any inactive rows, they are never used in the multiplication routine. Since the number of active rows in $B$ is equal to the number of active columns of $A$, the cost of loading $B$ can be approximated with $L_b |a_c|$, with $|a_c|$ representing the number of active columns in $A$.

The overall cost of SPMM with the LIBXSMM is given by: 
\begin{equation}
\label{eq:sparsepred}
	T = |a_r| * L_c +  nnz * L_a + |a_c| * L_b
\end{equation}


With an accurate estimation of $L_a$, $L_b$, and $L_c$, we can predict the execution time of a sparse-dense matrix multiplication just from the structure of the sparse matrix. Note that this structure is known \textit{a priori}, being the sparse matrices the pruned weights of the neural model. We begin by observing that $L_b$ and $L_c$ both describe memory operations, with the difference that $L_c$ measures both data reading and writing, while $L_b$ refers to data reading. We empirically verify that both the operations have the same time cost, \textit{i.e.}, $L_c = 2 L_b$. 

We now infer the coefficients $L_a, L_b, L_c$, starting from $L_b$. We cannot measure with a timer the cost of the elementary operations we have divided the LIBXSMM SPMM routine in, but we can empirically compute them by difference.

Let us consider two different sparse matrices $A_c$ and $A_{rd}$ with the same shape $m \times k$ and the same number of non-zero entries ($nnz$). $A_c$ has the non-zero values disposed on the same column $j*$, \textit{i.e.}, is a matrix where $a_{i,j} = 0 $ if $j \neq j*$. $A_{rd}$ is a sparse matrix which has single non-zero entry for each row and each column, \textit{i.e.,} $\sum_{i=0}^{i=m-1} a_{i,j} = 1, \forall j=0, \dots, m-1$  and $\sum_{i=0}^{i=k-1} a_{i,j} = 1,  \forall i=0, \dots, k-1$. 
The cost of multiplying $A_c$ and $A_{rd}$
with a dense matrix is given by:
\begin{align}
\nonumber
T (A_c) = m * L_c + nnz *L_a + 1 * L_b \\
\nonumber
T({A_{rd}}) = m * L_c + nnz *L_a + k * L_b
\end{align}
so, 
$$T(A_{rd}) - T({A_c}) =  (k-1) *L_b$$
We can experimentally measure  $T(A_{rd})$ and $T({A_c})$ and use them to compute $L_b$, since $k$ in known. 

To derive $L_a$, we use the same $A_c$ as before and a second matrix $A_{2c}$, having $2*nnz$ non-zero entries, organized along two columns. The cost for multiplying $A_{2c}$ with a dense matrix is given by 
$$T (A_{2c}) = m * L_c + 2*nnz *L_a + 2* L_b$$
Since $L_b$ can be derived using the previous expression, we can subtract $T(A_{2c})$ and $A_c$ and obtain $L_a$ as:
$$L_a = (T(A_{2c} )- T(A_c))/nnz -L_b$$
Our aim is to compute size-agnostic $L_a$, $L_b$ and  $L_c$. We set $M=K$ and vary them in $\{200, 300, 400, 500 \}$ and we experiment $N \in \{16, 32, 64\}$. $L_b$ and $L_c$ depends on $N$ (and so do $L_c$, which is computed doubling $L_b$), so we normalize diving by $N$. We observe that when $N \geq 128$, the value obtained for $L_a$ and $L_b$ diverge w.r.t. to smaller batch size. In fact,  larger $N$ values ($N \geq 128$) break the hypothesis of $B$ residing inside the cache during the whole multiplication, which is a fundamental assumption of our time predictor.  The definitive time predictor parameters are computed as an average of their value obtained with different shape configurations.
We demonstrate the validity of our sparse predictor in Table~\ref{table:est_vs_real_exec_t_sparse}, where we report the predicted and the real execution time needed to multiply the weights of the first layer of several neural models with a random input. We restrain our experiments to the sparsity range obtained with pruning on our neural architectures. As we will detail later, at these sparsity levels the time required for SDMM  is negligible w.r.t. to its dense counterpart. We also evidence that our time predictor is specific to matrix multiplication, hence can be essentially applied to fully-connected layers. Convolution or attention-based architectures have different properties that require a specific investigation. We leave these analyses for future works.

As we can see, the predictor is capable of correctly estimating the execution time of different models at high levels of sparsity, with a small error. Specifically, the predictor can fruitfully distinguish between matrix with the same shape but with different sparsity percentages; two examples are the $200 \times 136$ and the $100 \times 136$ instances in Table~\ref{table:est_vs_real_exec_t_sparse}. First, we observe that with sparsity percentage in the order of $1\%$, SDMM execution times can vary up to $30\%$. Second, the time predictor correctly reflects this peculiarity, thanks to the deep understanding of the routine details that stands behind its development.


\begin{table}[htb]

	\adjustbox{max width=\columnwidth}{
	\centering
	\begin{tabular}{lrrrrrrr}
		\toprule
		\multirow{3}{*}{Shape} &   	\multirow{3}{*}{Sparsity}& 
		 \multicolumn{6}{c}{SDMM Time ($\mu s$)} \\
		\cmidrule{3-8}
		& & \multicolumn{2}{c} {$N=16$}& \multicolumn{2}{c} {$N=32$} & \multicolumn{2}{c} {$N=64$} \\
		\cmidrule{3-8}
		 & & Real & Pred. & Real & Pred. & Real & Pred.\\		
		 \midrule
		 \multirow{2}{*}{400$\times$136} &  0.995 & 0.2 & 0.2 & 0.4& 0.4&   0.9 &          0.8 
		 \\
		  &  0.986 &  0.4& 0.4& 0.9& 0.8&         1.9 &          1.6 \\
		   \arrayrulecolor{black!30}\midrule

 	300$\times$136 &  0.985 &0.3 &0.3 & 0.7& 0.7&        1.6 &          1.4 \\
 	 	  \arrayrulecolor{black!30}\midrule

 	 	\multirow{2}{*}{200$\times$136} &  0.982 &   0.3& 0.3& 0.5 & 0.5 &        1.0&          1.0 \\
 	  		&  0.971 &       0.4& 0.3& 0.7&0.6 &  1.5 &          1.3 \\
 	  	\arrayrulecolor{black!30}\midrule

 	 	\multirow{2}{*}{100$\times$136} &   0.989 & 0.1 &0.1 & 0.2& 0.2 & 0.5 &          0.4 \\
          &  0.967 & 0.2&0.2 & 0.3& 0.4&      0.7 & 0.7 \\ 
         \arrayrulecolor{black!30}\midrule

 	 	50$\times$136  &  0.987 &0.1 	& 0.1& 0.1 & 0.1 & 	   0.2 & 		  0.2 \\
 	 	\arrayrulecolor{black}
 	 	\bottomrule
	\end{tabular}
	}
	
	\caption{Some examples of our sparse time predictor with different values of $N$.}
	\label{table:est_vs_real_exec_t_sparse}

\end{table}

\section{Neural Engineering}
\label{sec:neuraleng}
In Section~\ref{sec:introduction}, we claim that ensembles of regression trees consistently outperform, both in terms of effectiveness and efficiency, NNs trained with the method proposed by Cohen \textit{et al.}~\cite{cohen2018universal}, when documents are scored on CPU. In this section, we break down a methodology used to create efficient neural models for ranking that can compete with ensembles of tree-based ones.

\subsection{Approximation of an Ensembles of Trees}
\label{subsec:approxbetter}
We employ the methodology proposed by Cohen \textit{et al.}~\cite{cohen2018universal} to train neural models that approximate the scores of an ensemble of regression trees. This approach is effective since we use a powerful model, \textit{i.e.}, an ensemble of regression trees, and a profitable learning strategy, \textit{i.e.}, a \textit{listwise} approach, to extrinsic the structure of the actual underlying probability distribution. This facilitates the learning process of a simpler model, \textit{i.e.}, a shallow neural network. The idea is inherited from a deep learning compression technique named \textit{Knowledge Distillation}~\cite{bucilua2006model, ba2014deep,DBLP:journals/corr/HintonVD15} in which a small, production-oriented, network (\textit{student}) is trained to mimic the output of a large and effective network (\textit{teacher}).

To fully leverage the benefits of this technique, we train an ensemble of regression trees with the best performance on a validation set without taking into account its efficiency. Then, we use its scores as ground truth in a distillation process that trains our neural models. In Table~\ref{table:imprteacher}, we report the validity of this approach using the \msn dataset~\cite{DBLP:journals/corr/QinL13}, a widely adopted LtR dataset composed of more than 30,000 queries, with about 120 documents per query, where 
each document is a vector of 136 features\footnote{The list of features is available at https://www.microsoft.com/en-us/research/project/mslr/}. We adopt the NDCG@10 as quality metric. First, we observe the difference in terms of ranking precision between: 1) a model trained with a fixed number of leaves, \textit{i.e.}, $64$, 2) the best model we could obtain on the \msn dataset. The latter one, which results to have $256$ leaves per tree, consistently outperforms the $64$-leaves model.
Increasing the number of leaves in tree-based models allows for a remarkable gain in terms of NDCG@10. Indeed, when scoring a tree-based model with QuickScorer, the execution time scales at least linearly with the number of trees and leaves~\cite{lucchese2015quickscorer,dato2016fast}. Hence, a $256$-leaves model is more than $4$x slower than a $64$-leaves one with the same number of trees. 
In fact, given that the scoring time per document of $64$-leaves models is 8.2 $\mu s$, a $256$-leaves one takes at least $33 \mu s$ to be traversed with QuickScorer.
This means that, when pursuing a trade-off between effectiveness and efficiency, the best solution is the ensemble of $878$ trees with $64$ leaves, due to the linear dependency of the scoring time with respect to the number of leaves. 

Furthermore, we report the results when using two tree-based models as teachers for two different neural networks. 
Our experiments clearly show the positive effects of approximating a more effective \textit{teacher} (Table~\ref{table:imprteacher}). In fact, thanks to the teacher upgrade, the  $1000\times500\times500\times100$ can provide the same ranking precision as the $64$-leaves tree-based model. Observe that the \textit{student} is teacher-agnostic: the architecture of the network is independent w.r.t. the tree-based model which is approximating, and so is the time to perform the forward pass. In conclusion, distilling from a more effective teacher bridges the gap between neural models and ensemble of regression trees in terms of effectiveness. Nevertheless, a margin still exists between the two families of models in terms of efficiency. In the following sections, we will show how to tackle this aspect. 


\begin{table}
\centering
	\begin{tabular}{llr}
		\toprule
		Model  & Teacher & NDCG@10 \\
		\midrule	
		878 trees, 64 leaves &  / &  0.5246  \\
		600 trees, 256 leaves &  / &  $\uparrow$ \textbf{0.5291}     \\
		\midrule
		\multirow{2}{*}{500$\times$100} & 878 trees, 64 leaves & 0.5180 \\
		& 600 trees, 256 leaves &$\uparrow$ \textbf{0.5198} \\  
		\midrule
		\multirow{2}{*}{100$\times$500$\times$500$\times$100} & 878 trees, 64 leaves & 0.5208 \\
		& 600 trees, 256 leaves &$\uparrow$ \textbf{0.5243}  \\
		\bottomrule
	\end{tabular}
	\caption{Comparison in terms of NDCG@10 among Neural Networks on \msn, when trained to approximate different teachers.  $\uparrow$ indicates statistically significant improvement (Fisher's randomization test,  $p < 0.05$).  }
	\label{table:imprteacher}
\end{table}

\subsection{Design of a Neural Model}
\label{subsec::neuraldesign}
In this section, we present our novel methodology to design efficient neural models for ranking. We leverage the insights gained in studying dense and sparse matrix multiplication to show how to make correct architectural choices, thus training a very limited set of candidate models. We provide an empirical evaluation to show the correctness of our assumptions. 
Experiments are conducted on the  \msn dataset~\cite{DBLP:journals/corr/QinL13}, as in Section~\ref{subsec:approxbetter}. We first show how to develop dense models matching some given time requirements. Then, we employ pruning techniques to sparsify these models and outperform ensembles of regression trees.

\smallskip
\noindent \textbf{Architecture design}.
Our approach begins by choosing the dense architectures matching some given time constraints. For the sake of simplicity, we will assume to have two tree-based models to compete with, a $300$-trees ensemble and a $500$-trees ensemble, each one with $64$ leaves per tree. Their NDCG@10  and their scoring time ($\mu s$) are reported in Table~\ref{table:widevsdense}. By using the time predictor developed in Section~\ref{subsec:densetimepred}, the identification of the architectures matching the time requirements is now an easier task.
We build a heatmap as in Figure~\ref{fig:heatmap} and then use it to predict the execution time of the architecture, without the need of testing its performance on real hardware. This allows to discard models that do not match the desired latency constraints. As reported in Table~\ref{table:widevsdense}, there can be several models fitting the time budget. In our case, we propose $2,3,4$ layers NNs. We train the chosen models and compare their NDCG@10. \textit{Deep} networks (more layers) afford better performance w.r.t. \textit{wide} ones (more neurons per layer), coherently with the evolution of neural models witnessed in the last decade. The reason is that deep networks are generally capable of extracting higher levels features thus creating more complex representation of the input. The higher representations are built on simpler ones, generating a nested hierarchy of concepts which allows to improve the understanding and the learning from the data~\cite{Goodfellow-et-al-2016}. We empirically verify that $5$-layers models matching the time constraints do not offer advantages with respect to $4$-layers ones, showing that $4$-layer networks are expressive enough for the ranking task. Dense networks offer performance close to the tree-based model but do not really guarantee advantages neither in terms of effectiveness or efficiency, as shown in Table~\ref{table:widevsdense}.

\begin{table}
\centering
	\begin{tabular}{lrr}
		\toprule

		Model & Scoring Time ($\mu s$/doc) & NDCG@10 \\
		\midrule

		QuickScorer 300, 64 & 3.0  & 0.5230	\\
		\cdashlinelr{1-3}
		500$\times$100 & 2.2 & 0.5196 \\
		300$\times$200$\times$100 & 2.4 &  0.5209 \\
		300$\times$150$\times$150$\times$30 & 2.2 & 0.5207 \\
		\midrule
		QuickScorer 500, 64 & 4.9  & 0.5240	\\
		\cdashlinelr{1-3}
		1000$\times$200 & 5.5 & 0.5150 \\
		600$\times$300$\times$100 & 5.6 & 0.5203\\
		500$\times$250$\times$250$\times$100 & 5.4 & 0.5218\\
		\bottomrule
	\end{tabular}
	\caption{Comparison in terms of Scoring Time between QuickScorer and Neural Networks on \msn. The notation ''QuickScorer $x,y$'' indicates that $x$ is the number of trees, and $y$ the number of leaves per tree.   }
	\label{table:widevsdense}
\end{table}

\begin{figure}[t]
\begin{minipage}[b]{0.5\columnwidth}
\includegraphics[width=\columnwidth]{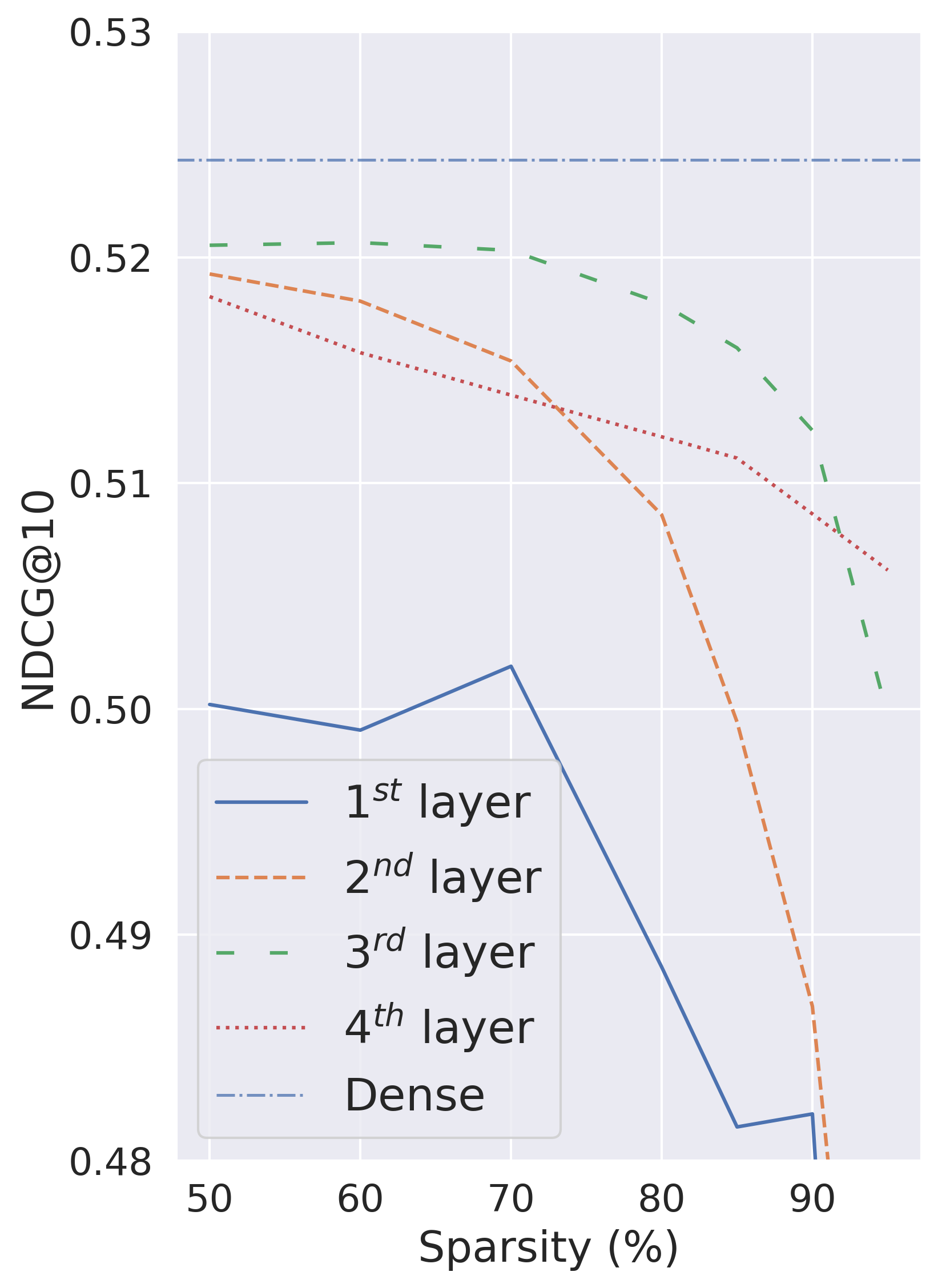}
\centering 
\caption*{\footnotesize{Static}}
\end{minipage}%
\begin{minipage}[b]{0.5\columnwidth}
\includegraphics[width=\columnwidth]{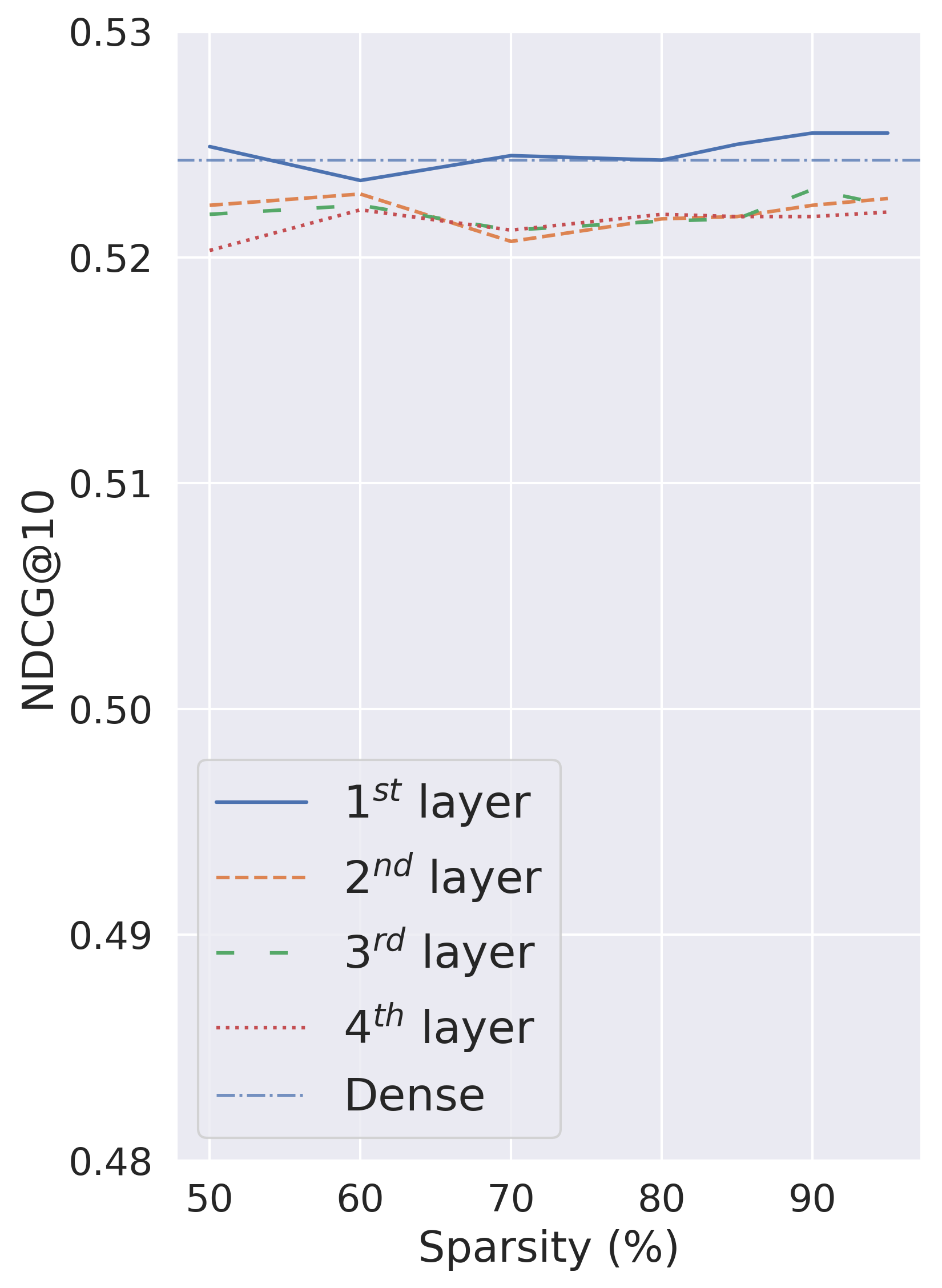}
\centering 
\caption*{\footnotesize{Dynamic}}
\end{minipage}%
\caption{Static and Dynamic Sensitivity Analysis for a 400$\times$200$\times$200$\times$100 network on the \msn dataset.}
\label{fig:sensitivity}
\end{figure}

\smallskip
\noindent \textbf{Sensitivity analysis and pruning.}
In our experiments, dense models do not reach the performance of ensembles of regression trees scored with QuickScorer. We now address the problem by leveraging the advantages brought by \emph{model compression}, in particular by \emph{network pruning} \cite{DBLP:journals/corr/HanPTD15,DBLP:journals/corr/GuoYC16}, a technique that deeply sparsifies a neural model without incurring in performance degradation. Let us consider the time budget of $3 \mu s $:
we devise a model which exceeds the time budget but affords an NDCG@10 close the $300$ trees model. By mean of pruning, we can move to the sparse domain and benefit of fast Sparse Dense Matrix Multiplication routines (SDMM). As example model, we pick a 400$\times$200$\times$200$\times$100 network: its performance are reported in Table~\ref{table:sparse_400x200x200x100_partial}.
As detailed in Section \ref{subsec:modelcompr}, \textit{magnitude-based} pruning methods deliver high compression rates without accuracy loss. We adopt this family of pruning techniques to sparsify the parameters of our model. Recall that magnitude pruning technique zero-out a given amount of low absolute value weights. The amount of zeroed-out values determines the aggressiveness of the sparsification: the way this aggressiveness is controlled distinguishes between level pruning and threshold-based pruning.
In the case of level pruning, we can explicitly set the sparsity target, \emph{e.g.}, 70\%.
In the threshold-based magnitude pruning by Han \textit{et al.}~\cite{DBLP:journals/corr/HanPTD15}, instead, we need to chose a statistical based threshold, as detailed in Section~\ref{sec:related}. The choice is generally based on the \textit{sensitivity} of each layer, namely  the property that describes a layer's resistance to sparsification.
We perform two kind of sensitivity analysis: \textit{static} and \textit{dynamic}. Both procedures prune a growing percentage of weights in each layer, one layer at a time, and evaluate the behavior of the partially-pruned model on the validation set. In the static version, there is no re-training~\cite{DBLP:journals/corr/HanPTD15} of the weights that survived the pruning in the chosen layer and the weights in the other layers, while in the dynamic version re-training is applied.
Static analysis is reported on the left side of Figure~\ref{fig:sensitivity}. The sensitivity of each layer appears to decrease as we go deeper into the network, meaning that the first layers suffer the most from sparsification. Dynamic analysis (right side of Figure~\ref{fig:sensitivity}) shows an inverse trend and highlights a peculiar behavior of the first layer: high levels of sparsity in this layer allow the pruned model to outperform the dense one in terms of NDCG@10.  This is an example of a model compression technique acting as regularizer \cite{DBLP:journals/corr/HanPTD15,DBLP:journals/corr/ZhouYGXC17}.
T
This effect is especially evident in the first layer as it presents the weights with the largest absolute values among all the other network layers. Observe that the effect of matrix multiplication is dominated by large absolute value entries, and the larger are the values, the larger is their impact. So a reduced number of high absolute weights can well approximate the overall result of matrix multiplication. From a learning point of view, since the network is working on handcrafted features, the sparsification selects just the essential combinations of input features.

\begin{figure}[t]
\centering
\includegraphics[width=\columnwidth]{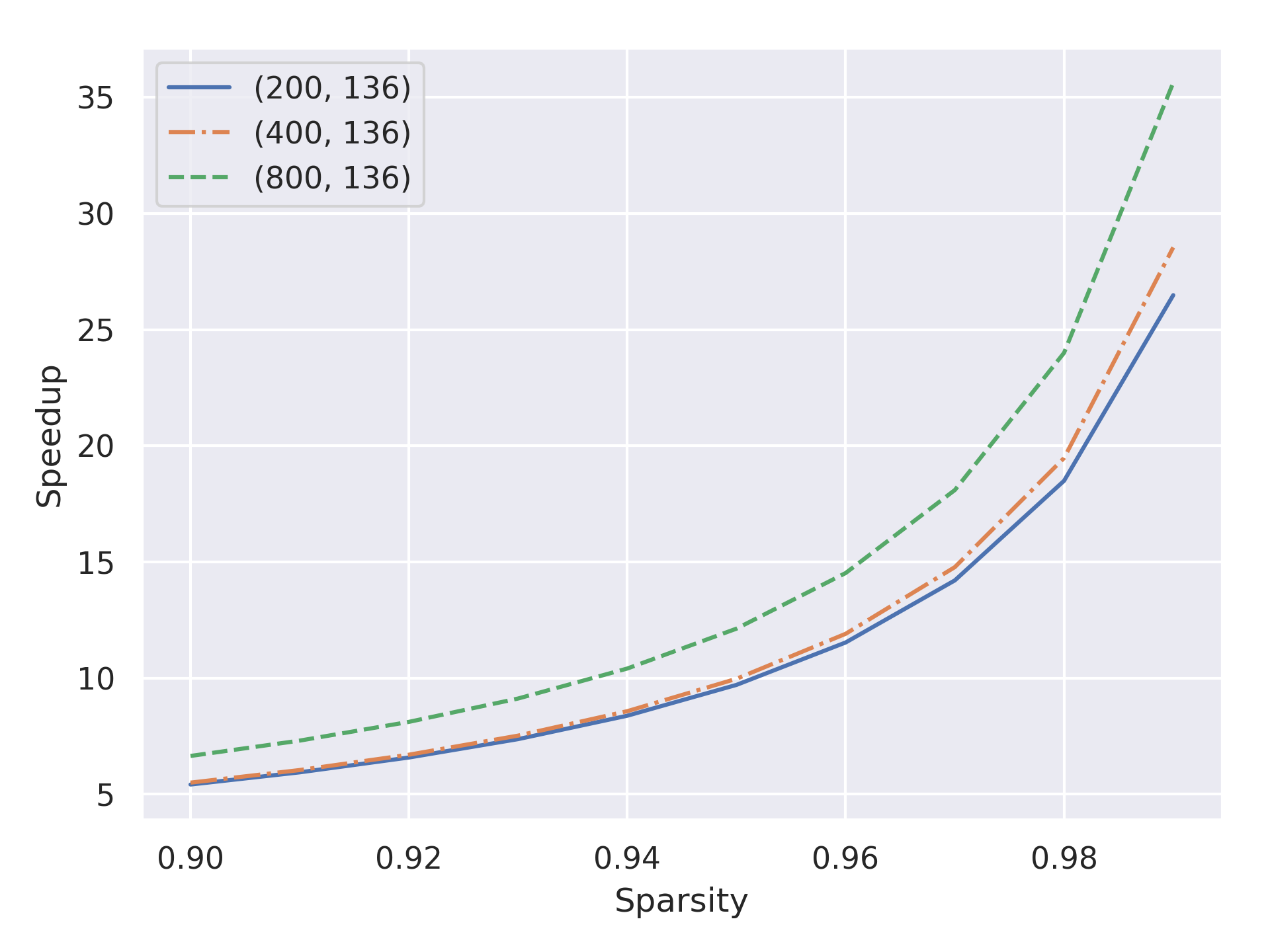}
\caption{Matrix multiplication speedup at various levels of sparsity estimated with our sparse time predictor. We assume the number of active columns/rows to be equal to the total number of columns/rows (worst-case scenario).\label{fig:sparsespeedup}}
\end{figure}

\begin{table}[b]
	\centering
	\adjustbox{max width=\columnwidth}{
	\begin{tabular}{lR{0.6cm}R{0.6cm}R{0.6cm}R{0.6cm}R{0.6cm}}
		\toprule
		\multirow{2}{*}{Model} & \multicolumn{5}{c}{\footnotesize{Relative Execution Time per Layer (\%)}} \\
		\cmidrule{2-6}
		 & \nth{1} & \nth{2}& \nth{3} &\nth{4}&\nth{5} \\
		\midrule
		400$\times$200$\times$200$\times$100 & \textbf{35} & 33 & 20 & 10 & 2   \\
		100$\times$50$\times$50$\times$10 & \textbf{60} & 21 & 14 & 3 & 2 \\
		200$\times$100$\times$100$\times$50 & \textbf{45} & 28 & 17 & 8 & 2 \\
		\bottomrule
	\end{tabular}
	 }
	\caption{Breakdown of the relative execution time among different layers for different neural models. }
	\label{table:breakdown}
\end{table}
Pruning techniques were originally developed to reduce the size of pre-trained models~\cite{DBLP:journals/corr/HanPTD15,DBLP:journals/corr/GuoYC16}. 
Despite, in our context we aim at speeding up the forward pass without incurring in performance degradation. 
This induces us to consider each layer's relative impact on the inference step before applying a pruning technique. In Table~\ref{table:breakdown}, we report a breakdown of the execution times among different layers in different architectures. Observe that the most time-consuming layer is always the first one, even if the largest matrix is the one storing the second layer weights, as for the 400$\times$200$\times$200$\times$100 network. 
Applying bias and ReLU6, in fact, causes the output matrix of the first layer to be brought into the cache, where it resides there during the computation of the second layer. 
Observe also that it is sufficient to reduce the execution time of one of the first two layers to match the time budget  of $3 \mu s$. By using our sparse time predictor we can infer the required sparsity to obtain a given speedup. In Figure~\ref{fig:sparsespeedup}, we draw the sparsity-speedup curve for some matrices, representing the first layers of different architectures. Even if the dense first layer usually has a major impact on the overall execution time, the quadratic growth of the sparse speedup in the selected range annihilates its contribution after the sparsification. For example, in the 400$\times$200$\times$200$\times$100 architecture, the impact of the first layer in the dense version is about $35\%$, while at $95\%$ of sparsity, the estimated speedup using sparse multiplication is $10$x, meaning that the first layer after pruning becomes the second less time-consuming layer after \textit{fc5}.

\smallskip
\noindent \textbf{Outperforming tree-based models}.
By jointly harvesting 1) the prominent impact of the first layer on the total execution time, and 2) the regularization effect of pruning the first layer, we develop our \textit{ early-layers efficiency-oriented pruning}. We apply the threshold-based magnitude pruning using the Distiller framework~\cite{nzmora2019distiller}, a deep learning compression framework developed by Intel. This pruning technique generally offers more flexibility and better performance with respect to level pruning. We prune only the first layer in an aggressive fashion and we fine-tune its surviving entries and all the weights of the other layers.
In our final model, the first layer is $98.7\%$ sparse, meaning that there are about $700$ surviving non-zero weights in the first layer, out of 54400 ($400 \times 136$) in the dense matrix.
We use our sparse time predictor to compute the execution time. The speedup obtained with this sparsity ratio on the multiplication of the first layer is about $25$x. This means that the impact of the first layer, which previously amounted to about the $35\%$, is negligible.
In Table~\ref{table:sparse_400x200x200x100_partial} we report the comparison between tree-based models and neural models. While the dense model did not offer any advantages with respect to the tree-based models, the hybrid model - first layer sparse, other layers dense - is both the fastest and the most accurate model. For example, at the same NDCG@10 value, it is $3.2$x faster than the $878$-trees model.

\begin{table}[t]
	\centering
	\begin{tabular}{llrr}
		\toprule
		Model & Description & NDCG@10 & Sc. Time ($\mu s$/doc) \\
		\midrule
		\multirow{3}{*}{QuickScorer} & 878 trees & $\uparrow$ \textbf{0.5246} & 8.2 \\
		& 500 trees & 0.5240 & 4.9 \\
		& 300 trees & 0.5230 & 3.0 \\
		\midrule
		\multirow{2}{*}{Neural} & Dense & 0.5222 & 3.8 \\
		& Sparse & $\uparrow$ \textbf{0.5246} & \textbf{2.6} \\ 
 		\bottomrule
	\end{tabular}
	\caption{Dense and sparse neural models (400$\times$200$\times$200$\times$100)  vs QuickScorer in terms of NDCG@10 and Scoring Time (Sc. Time). $\uparrow$ indicates statistically significant improvement w.r.t. models of the same family (Fisher's randomization test,  $p < 0.05$).\label{table:sparse_400x200x200x100_partial}}
\end{table}

%% file: 04.experiments.tex

\section{Experiments}
\label{sec:experiments}
In this section, we provide an extensive evaluation of our methodology to design, train and sparsify neural models for the document scoring task. In particular, we compare them against tree-based models at different points of the efficiency-effectiveness trade-off. Throughout this article, we have used the \msn dataset as use case. We now complement our evaluation with the \istella dataset~\cite{dato2016fast}. 
First, we present the experimental setup. Then, we report our experimental results and we show that neural models obtained with our technique can outperform ensembles of trees. To ease the reproducibility of the results presented in this article, code and trained models have been made publicly available\footnote{\url{https://github.com/hpclab/efficient_nn_for_ltr}}.

\vspace{-.3cm}
\subsection{Experimental Setup}
\label{subsec:expsetup}
We perform our experiments on two datasets: \istella and \msn. The \istella dataset~\cite{dato2016fast} consists of a collection of $33$,$018$ queries with an average of $103$ documents per query. Each document-query pair is represented by $220$ features. The \msn (Fold 1) dataset, which we already introduced, is composed by more than $30$,$000$ queries, with about $120$ documents per query and 136 features per document-query pair. In both the dataset, document-query pairs are labeled with $5$-graded relevance judgments ranging from 0 (irrelevant) to 4 (perfectly relevant).
Both datasets are split in train-validation-test according to a 60\%-20\%-20\% criterion. 

\smallskip
\noindent \textbf{LambdaMART models}. We employ the LightGBM framework~\cite{NIPS2017_6907} to train ensembles of regression trees using the LambdaMART algorithm. For each training, we perform hyper-parameter tuning using the HyperOpt library \cite{bergstra2013making}.
In particular, we determine the optimal combination of the following set of hyper-parameters: \texttt{learning rate, max depth, min\_sum\_hessian\_in\_leaf, min\_data\_in\_leaf}. 
To avoid overfitting, we apply an early stopping criterion on the validation loss every 100 trees. We train 64-leaves model as target model to compare against neural networks and 256-leaves models to use as teachers. The latter models offer higher retrieval performance while being $4$x slower, which is not suitable for the use in latency-bounded applications. We score the LambdaMART models using a C++ implementation of QuickScorer that exploits instruction-level parallelism by using AVX2 instructions~\cite{8035185}.

\smallskip
\noindent \textbf{Neural Networks}. We train neural models (\textit{students}) to approximate the scores of top-performing regression forest (\textit{teacher}), accordingly to the knowledge distillation \cite{ba2014deep} paradigm, detailed in Section~\ref{subsec:approxbetter}. Models are trained using Pytorch~\cite{NEURIPS2019_9015}, adopting the same strategy for randomly generating training data of Cohen \textit{et al}.~\cite{cohen2018universal}. We employ RELU6 as activation function after every linear layer, except for the last one, where $\text{RELU6}(x) = min(max(x,0), 6)$. 
We use the Distiller~\cite{nzmora2019distiller} framework to prune the neural networks. Both in training and pruning, we employ Adam~\cite{kingma2014adam} as optimizer, with learning rate $0.001$ and no weight decay. Table~\ref{table:neuraltrainparams} summarizes the other training and pruning hyper-parameters, which are dataset-dependent. $E_t$ represents the number of training epochs. The pruning phase is composed of $E_p$ epochs of pruning/fine-tuning and of $E_{ft}$ epochs of only fine-tuning, as done by Han \textit{et al.
\cite{DBLP:journals/corr/HanPTD15}}.
Both for training and pruning, we scale the learning rate by multiplying it by $\gamma$ at the epochs specified by $\gamma_{step}$. Dropout, if employed (see Table~\ref{table:neuraltrainparams}), is applied only after the first layer.
When training and pruning the neural models, we always distill from the most effective ensemble of regression trees for the current dataset. On \msn, it is a model with 600 trees and 256 leaves per tree, reaching 0.5291 of NDCG@10, while on \istella it is a forest with $2500$ trees with $256$ leaves per tree, reaching 0.7821 of NDCG@10.	
 The neural forward pass is implemented in C++. We use the \textit{dnnl\_sgemm} routine from the OneDNN framework for dense matrix multiplication and the LIBXSMM~\cite{heinecke2016libxsmm} C++ library for sparse-dense matrix multiplication (after pruning).

\begin{table}[htb]
	\centering
	\begin{tabular}{lrrrrrr}
		\toprule
		Dataset & $E_t$ & $E_p$& $E_{ft}$& $\gamma$& $\gamma_{step}$ & Dropout \\
		\midrule
		\msn & 100 & 80 & 20 & 0.1 & $ 50, 80 $ & - \\
		\istella & 250 & 60 & 190 & 0.5 & $ 90, 130, 180 $ & 0.1 \\
		\bottomrule
	\end{tabular}
	\caption{Training and pruning parameters employed for neural networks on \msn and \istella.\label{table:neuraltrainparams}}
\end{table}

\smallskip	

\noindent \textbf{Experimental Methodology}. We experimentally evaluate the performance of neural networks and ensemble of regression trees on two different experimental scenarios:

\begin{itemize}
	\item \textit{High-Quality Retrieval}: this scenario covers use cases where high-precision retrieval is required, even at the price of a larger scoring time. We impose a constraint on the retrieval quality to our models, specified by a threshold on the ranking metric. As threshold, we choose the 99\% of the retrieval quality of the top performing tree-based competitor on each dataset.
	\item \textit{Low-Latency Retrieval}: this scenario is orthogonal to the previous one as it focus on the efficiency of the retrieval process. We specify a maximum per-document scoring time and we select only the models that can match it. For both datasets, we set the maximum per-document scoring time to be $0.5\mu s$.
\end{itemize}
We perform the comparison between neural models and ensemble of regression trees by considering one scenario at a time. For each dataset, we consider the Pareto frontier of ensembles of tree-based models respecting the constraint of the considered scenario (green lines in Figures~\ref{fig:hq},~\ref{fig:lowlat}). By doing so, we train several tree-based competitors at different efficiency-effectiveness trade-offs. We then apply our technique and we show that neural networks can outperform ensembles of regression trees. We employ our time predictors to train and prune only neural network models that fit the time budget constrained by the ensembles of tree-based models considered. We recall that our methodology allows to train a neural model and to prune its first layer. In fact, in Section~\ref{sec:neuraleng}, we demonstrate that the first layer has a prominent impact on the overall execution time. By zeroing out at least  95\% of the parameters, its impact becomes negligible (Figure~\ref{fig:sparsespeedup}). Furthermore, the sparsification of the first layer has a positive effect on the generalization capabilities of the model as it act as a regularizer.
Then, we forecast the overall execution time by subtracting the contribution of the dense first layer from the overall execution time. Both times can be estimated with our dense predictor with no computational effort.


\smallskip
\noindent \textbf{Experimental Platform}. 
All the inference algorithms are compiled with GCC 9.2.1 with the \texttt{-O3} compiler option. 
Scoring times are measured on a Intel i9-9900K CPU clocked at 3.5 GHz, with AVX2 instructions, with a L1-cache of 256KiB, a L2-cache of 2 MiB, and a L3-cache of 16MiB. All the scoring experiments have been performed in single-thread execution.

\vspace{-0.2cm}
\subsection{Results}

\begin{figure}[t]
\begin{minipage}[b]{0.5\columnwidth}
\includegraphics[width=\columnwidth]{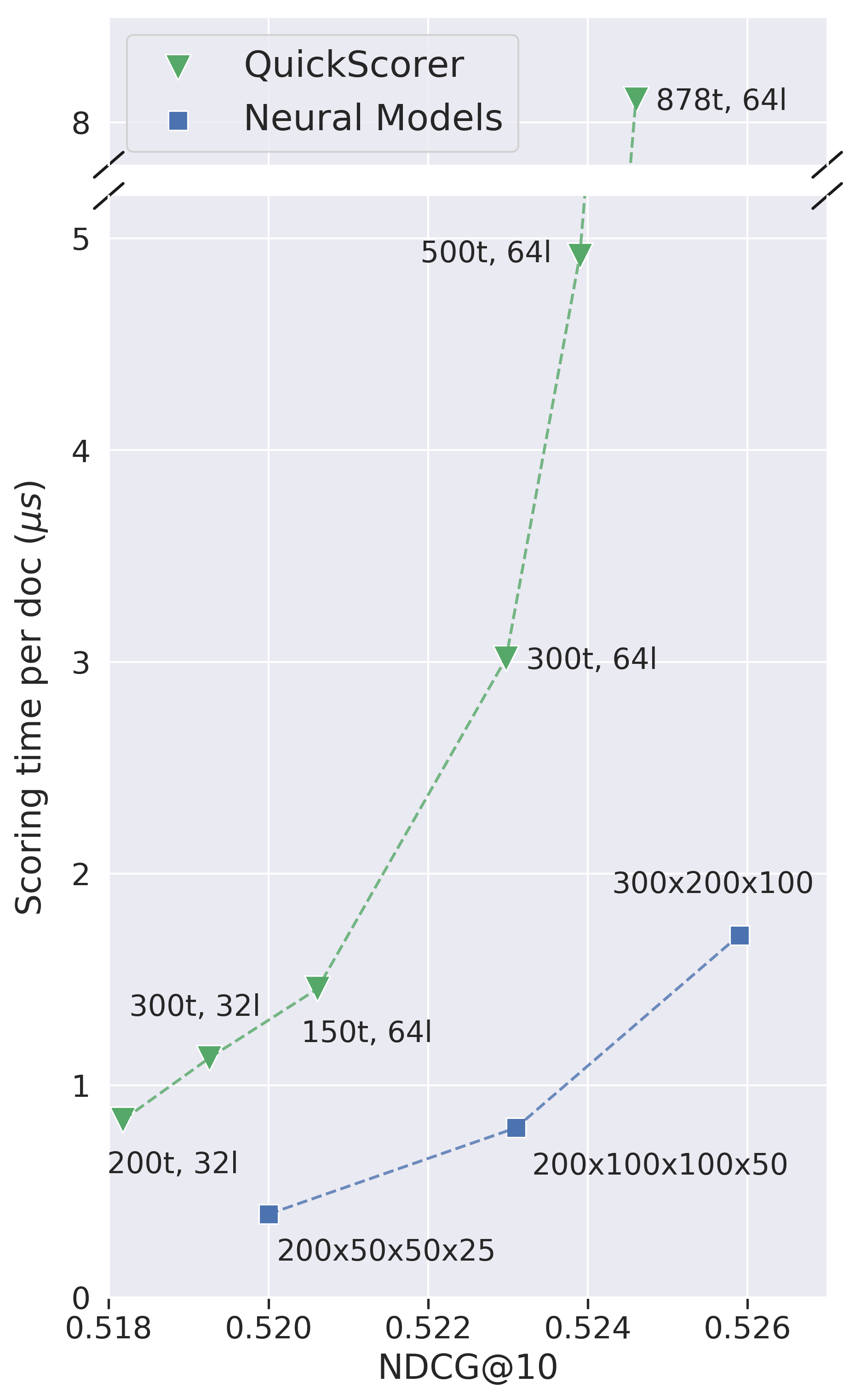}
\caption*{\footnotesize{\msn}}
\end{minipage}%
\begin{minipage}[b]{0.517\columnwidth}
\includegraphics[width=\columnwidth]{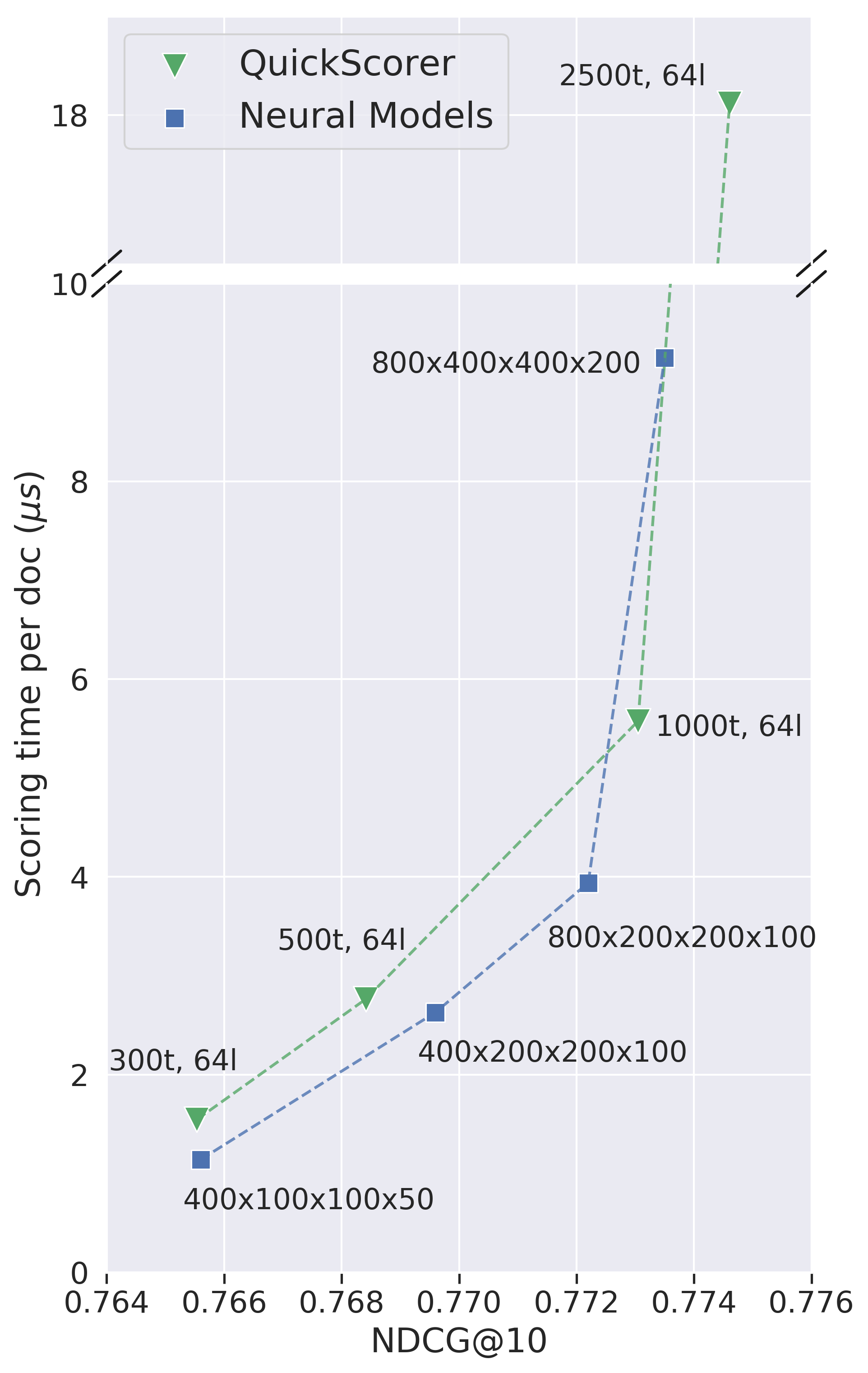}
\caption*{\footnotesize{\istella}}
\end{minipage}%
\caption{Comparison between neural networks and ensemble of regression tree on the \textit{high-quality retrieval} scenario.\label{fig:hq}}
\end{figure}
\noindent \textbf{High-Quality Retrieval}.
The first scenario of our comparison involves models delivering high-quality ranking. As previously detailed, we consider a model (both neural and tree-based) to be in the high-quality ranking region if its NDCG@10 is at least the 99\% of the top-quality tree model with 64 leaves.
By following the experimental methodology described above, we first construct the Pareto frontier for the ensemble of regression trees (green line in Figure\ref{fig:hq}).
We then move to the design of the neural network models. We can estimate the execution time of a neural model whose first layer is sparse with our time predictors. In particular, in Table~\ref{table:msn_final} we report the estimated execution time for the dense architecture, the relative impact of the first layer on the overall execution time, and the predicted execution time after pruning the first layer.
We always assume the sparsity of the first layer in the final model to be above $95\%$, so that its impact on the overall execution time is negligible. Our experiments show that this level of sparsity does not hamper the ranking capability of the model.
Observe that our time predictors permit to locate a neural model on the $y$-axis of the effectiveness-efficiency plot without any computational effort, analytically computing it given the architectures of network. 
Once we have designed our models to compete with the tree-based ones, we train and prune them, according to the methodology listed in Section~\ref{subsec:expsetup}.
\begin{table}[b]
	\centering
	\adjustbox{max width=\columnwidth}{
	\begin{tabular}{llrrrr}
		\toprule
		\multirow{2}{*}{Dataset} & \multirow{2}{*}{Model} & Sc. Time  & \nth{1} layer & Predicted Pruned   \\
		&& ($\mu s$/doc) & impact (\%)& Sc. Time  ($\mu s$/doc) \\
		\midrule	
		\multirow{3}{*}{\msn} & 300$\times$200$\times$100 & 2.4 & 30 &  1.7 \\
		&\footnotesize 200$\times$100$\times$100$\times$50 & 1.3 & 39 & 0.8 \\
		&200$\times$50$\times$50$\times$25 & 0.9 & 58 & 0.4\\
 		\midrule	
 		\multirow{3}{*}{\istella} & \footnotesize800$\times$400$\times$400$\times$200& 11.9 & 23 &  9.1 \\
		&\footnotesize800$\times$200$\times$200$\times$100	& 6.5 & 41 & 	3.8 \\
		&300$\times$200$\times$100 & 2.8 & 41 & 1.6	\\
		\bottomrule
	\end{tabular}}
	\caption{Prediction of model scoring time (Sc. Time) when pruning the first layer, in \emph{High Quality Retrieval}.  }
	\label{table:msn_final}
\end{table}

Figure~\ref{fig:hq} illustrates the comparison on a effectiveness-efficiency plot between neural models and ensemble of regression trees scored with QuickScorer. On the $x$-axis we report the NDCG@10 on the test set, and on the $y$-axis the scoring time per document in $\mu s$. First, we observe that the predicted times reported in Table~\ref{table:msn_final} coincide with real scoring time, confirming the precision of our theoretical approach. Hence, our methodology allows to train exclusively the required architectures. Secondly, neural models can outperform tree-based models in scoring documents, both in terms of effectiveness and efficiency. The neural Pareto-optimality, reported in blue in  Figure~\ref{fig:hq}, lays below the tree-based one (in green), either on the \msn dataset and on \istella. On the \msn dataset, for example, the 300$\times$200$\times$100 architecture is $4.4$x faster than the $878$-trees model and it also provides a higher retrieval quality. Furthermore, the 200$\times$50$\times$50$\times$25 architecture is the fastest model respecting the quality constraint on this dataset. The same consideration holds for \istella, where the fastest model respecting the imposed quality constraint is a neural network (400$\times$200$\times$200$\times$100).
On this dataset, neural models still outperform ensemble of regression trees on a large portion of the selected effectiveness-efficiency space, even if tree-based model deliver a slightly superior performance in the top performing region.
This leaves space for research work to further improve the quality of this approximation. 

\begin{figure}[t]
\begin{minipage}[b]{0.5\columnwidth}
\includegraphics[width=\columnwidth]{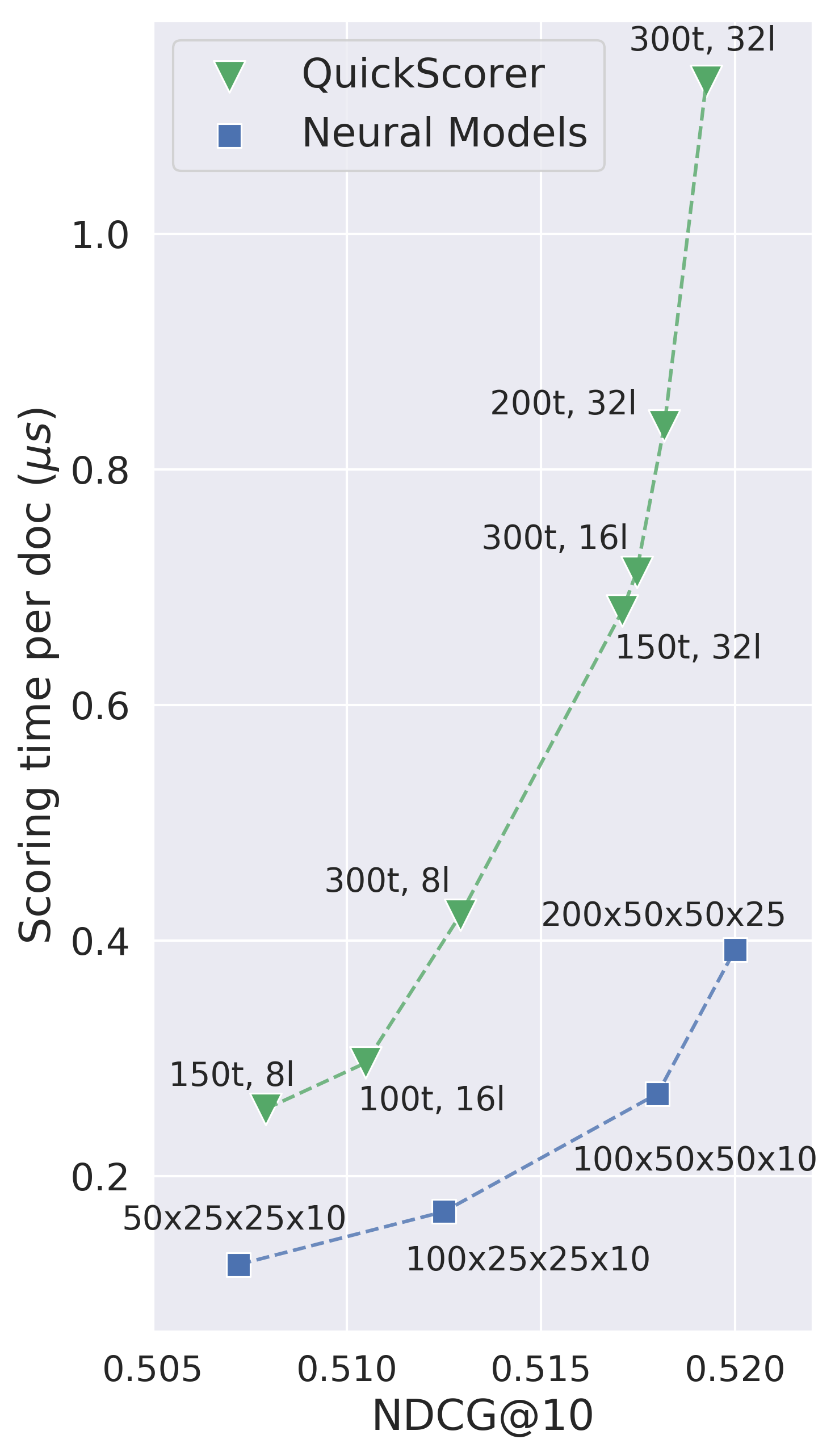}
\centering 
\caption*{\footnotesize{\msn}}
\end{minipage}%
\begin{minipage}[b]{0.51\columnwidth}
\includegraphics[width=\columnwidth]{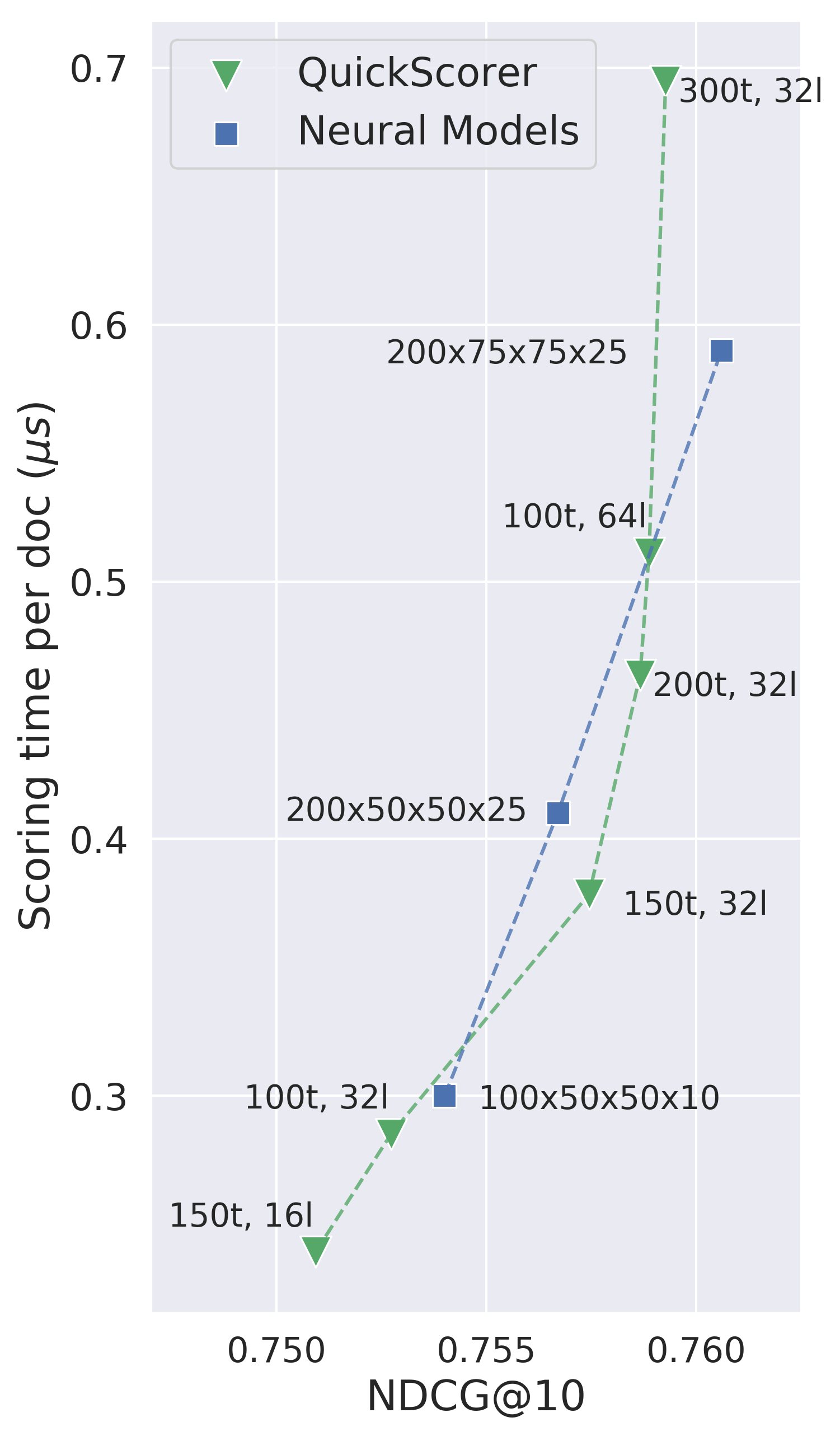}
\centering 
\caption*{\footnotesize{\istella}}
\end{minipage}%
\caption{Comparison between neural networks and ensemble of regression tree on the \textit{low-latency retrieval} scenario.}
\label{fig:lowlat}
\end{figure}

\smallskip
\noindent \textbf{Low-Latency Retrieval}. We now compare neural models and ensemble of regression trees on a low-latency retrieval setting, \textit{i.e.}, a scenario requiring the scoring time to be lower than $0.5 \mu s$ per document. The Pareto Optimality curve for the ensemble of regression trees is drawn in green in Figure~\ref{fig:lowlat}. We use this plot to identify the latency constraints for our neural networks. Our proposed methodology permits to precisely estimate the execution time of a model, before carrying out the costly training-pruning phase. In Table~\ref{table:msn_lowq}, we demonstrate the usage of our methodology. As we did for the \emph{high-quality retrieval} use case (Table~\ref{table:msn_final}), we report the predicted execution time for the dense architecture, the relative impact of the first layer and the predicted time after sparsification. We still consider the impact of the first layer to be negligible.
\begin{table}[b]
	\centering
	\adjustbox{max width=\columnwidth}{
	\begin{tabular}{llrrrr}
		\toprule
		\multirow{2}{*}{Dataset} & \multirow{2}{*}{Model} & Sc. Time  & \nth{1} layer   & Predicted Pruned  \\
		&& ($\mu s$/doc) & impact (\%)& Sc. Time  ($\mu s$/doc) \\
			\midrule
		\multirow{3}{*}{\msn}&100$\times$50$\times$50$\times$25 & 0.6 & 56 & 0.3\\
		&100$\times$25$\times$25$\times$10 & 0.5 & 71 & 0.2\\
		&50$\times$25$\times$25$\times$10 & 0.3 & 65 & 0.1 \\
 		\midrule
		\multirow{3}{*}{\istella}&  200$\times$75$\times$75$\times$25&1.6 & 61 &  0.6 \\
		&100$\times$75$\times$75$\times$10& 0.9 & 55 & 	0.4 \\
 		&100$\times$50$\times$50$\times$10 & 0.8 & 67 & 0.3	\\
		\bottomrule
	\end{tabular}
	}
	\caption{Prediction of model scoring time (Sc. Time) when pruning the first layer, in \emph{Low-Latency Retrieval}.\label{table:msn_lowq}}
\end{table}

Figure~\ref{fig:lowlat} illustrates the comparison between neural model and ensemble of regression trees when dealing with low-latency constraints. Even in this case, our methodology permits to create neural networks that  outperform ensembles of regression trees. On the \msn dataset, neural models dominate over tree-based models, as happened for the \textit{high-quality} use case. In fact,  the Pareto frontier of neural models (in blue) always lies below the tree-based one, confirming the superiority of our technique on this dataset (left side of Figure~\ref{fig:lowlat}). In particular, the 200$\times$50$\times$50$\times$25 architecture is $3$x faster than the regression forest with $300$ trees and $32$ leaves, while being also more precise in terms of NDCG@10. On the \istella dataset (right side of Figure~\ref{fig:lowlat}, the performance of our neural models can be considered on pair with tree-based models. In fact, the two Pareto frontiers intersect in this portion of the efficiency-effectiveness trade-off. Despite that, neural networks  still provide the most effective model respecting the time requirement (200$\times$75$\times$75$\times$25). 
This dataset confirms to be troublesome for neural models, as witnessed in the high-quality retrieval scenario. 

%% file: 05.conclusions.tex
\section{Conclusions and Future Work}
\label{sec:conclusions}
In this paper, we presented an effective and efficient methodology to design neural networks for document scoring in a modern information retrieval system. The neural models we take into account are trained to approximate the scores of an ensemble of regression trees. By leveraging a combination of high-performance dense-dense, sparse-dense matrix multiplication, and element-wise pruning, the neural models can compete with the original models.
Thus, our methodology is \textit{effective}. By developing time predictors based on an accurate study of how these operations are implemented on modern processors, we are capable to precisely estimate the execution time of a given architecture by knowing the shape and the sparsity level of each layer. This allows to train only a limited number of models, the ones matching the time requirements given by the specific context. Our methodology is thus \textit{efficient}. Besides presenting our method, throughout the paper emerges a comparison between ensembles of regression trees and NNs on the document scoring task, tested on the \msn and \istella datasets. In our experiments,  neural networks are not capable of reaching the accuracy of their \textit{teacher}, hence tree-based methods are superior in top-quality retrieval scenarios. At any other level of the efficiency-effectiveness trade-off, neural models designed and trained with our approach can always outscore or at least compete with ensembles of regression trees.

As future work, we intend to apply different compression methods such as quantization or early exiting to further improve the efficiency of our neural models. Moreover, we plan to extend our comparison between neural networks and ensemble of regression trees to other computational engines, such as General-Purpose Graphic Processing Unit (GPU) or Field Programmable Gate Array (FPGA). We also aim at improving the training by distillation procedure of neural models, in order to bridge the effectiveness gap with ensembles of regression trees. 

\smallskip
\noindent \textbf{Acknowledgements}.
This paper is partially supported by the ``Algorithms, Data Structures and Combinatorics for Machine Learning'' (MIUR-PRIN 2017) and the OK-INSAID (MIUR-PON 2018, grant agreement ARS01\_00917) projects.